\documentclass[final,5p,times,twocolumn,authoryear]{elsarticle}

\usepackage{tabularx}
\usepackage{array}
\newcolumntype{Y}{>{\centering\arraybackslash}X}

\usepackage{subcaption}
\usepackage{setspace}
\usepackage[inkscapelatex=false]{svg}
\usepackage{geometry}
\usepackage{epstopdf}
\usepackage{graphicx}
\usepackage{float}
\usepackage[labelsep=period]{caption}
\usepackage[british]{babel} 
\usepackage[hang]{footmisc}
\usepackage{amssymb}
\usepackage{lipsum}
\usepackage{enumitem}
\usepackage{amsmath}
\usepackage{eucal}
\usepackage{booktabs}
\usepackage{array}
\usepackage{url}
\usepackage[left]{lineno}
\usepackage{xspace}
\usepackage{multirow}
\usepackage{bm}
\usepackage{dsfont}
\usepackage{soul}
\soulregister\cite7
\soulregister\citep7
\soulregister\ref7
\soulregister\eqref7
\sethlcolor{yellow}
\usepackage{xcolor}
\usepackage[table]{xcolor}
\usepackage{algorithm}
\usepackage{algorithmic}
\usepackage{amssymb} 
\usepackage{pifont}
\usepackage{lineno}

\usepackage[colorlinks,citecolor=black,linkcolor=red]{hyperref}

\def\OurMethod{{TasselNetV4}\xspace}

\begin{document}


\begin{frontmatter}

\title{TasselNetV4: A vision foundation model for cross-scene, cross-scale, and cross-species plant counting}

\author[hust-aia]{Xiaonan Hu}

\author[hust-jl]{Xuebing Li}

\author[hust-cyberse]{Jinyu Xu}

\author[hust-aia]{Abdulkadir Duran Adan}

\author[hust-aia]{Letian Zhou}

\author[hust-aia]{Xuhui Zhu}

\author[wit]{Yanan Li}

\author[tokyo]{Wei Guo}

\author[nau]{Shouyang Liu}

\author[hust-jl]{Wenzhong Liu}

\author[hust-aia]{Hao Lu\corref{cor1}}

\cortext[cor1]{Corresponding author}
            
\affiliation[hust-aia]{organization={National Key Laboratory of Multispectral Information Intelligent Processing Technology, School of Artificial Intelligence and Automation}, 
            addressline={Huazhong University of Science and Technology}, 
            city={Wuhan},
            postcode={430074},
            country={China}}

\affiliation[hust-jl]{organization={China-Poland Joint Laboratory on Measurement and Control Technology, School of Artificial Intelligence and Automation},
            addressline={Huazhong University of Science and Technology}, 
            city={Wuhan},
            postcode={430074},
            country={China}}

\affiliation[hust-cyberse]{organization={School of Cyber Science and Engineering}, 
            addressline={Huazhong University of Science and Technology}, 
            city={Wuhan},
            postcode={430074},
            country={China}}

\affiliation[wit]{organization={School of Computer Science and Engineering}, 
addressline={Wuhan Institute of Technology}, 
city={Wuhan},
postcode={450205},
country={China}}

\affiliation[tokyo]{organization={Graduate School of Agricultural and Life Sciences},
            addressline={The University of Tokyo}, 
            city={1-1-1 Midori-cho, Nishitokyo City, Tokyo},
            country={Japan}}

\affiliation[nau]{organization={Engineering Research Center of Plant Phenotyping, Ministry of Education, Jiangsu Collaborative Innovation Center for Modern Crop Production, Academy for Advanced Interdisciplinary Studies}, 
            addressline={Nanjing Agricultural University}, 
            city={Nanjing},
            postcode={210095},
            country={China}}

\begin{abstract}

Accurate plant counting provides valuable information for agriculture such as crop yield prediction, plant density assessment, and phenotype quantification. Vision-based approaches are currently the mainstream solution. Prior art typically uses a detection or a regression model to count a specific plant. However, plants have biodiversity, and new cultivars are increasingly bred each year. It is almost impossible to exhaust and build all species-dependent counting models. Inspired by class-agnostic counting (CAC) in computer vision, we argue that it is time to rethink the problem formulation of plant counting, from what plants to count to how to count plants.
In contrast to most daily objects with spatial and temporal invariance, plants are dynamic, changing with time and space. Their non-rigid structure often leads to worse performance than counting rigid instances like heads and cars such that current CAC and open-world detection models are suboptimal to count plants. In this work, we inherit the vein of the TasselNet plant counting model and introduce a new extension, TasselNetV4, shifting from species-specific counting to cross-species counting. TasselNetV4 marries the local counting idea of TasselNet with the extract-and-match paradigm in CAC. It builds upon a plain vision transformer and incorporates novel multi-branch box-aware local counters used to enhance cross-scale robustness. In particular, two challenging datasets, PAC-$105$ and PAC-Somalia, are harvested. 
PAC-$105$ features $105$ plant- and organ-level categories from $64$ plant species, spanning various scenes. 
PAC-Somalia, specific to out-of-distribution validation, features $32$ unique plant species in Somalia. 
Extensive experiments against state-of-the-art CAC models show that TasselNetV4 achieves not only superior counting performance but also high efficiency, with a mean absolute error of $16.04$, an $R^2$ of $0.92$, and up to $121$ FPS inference speed on images of $384\times384$ resolution.
Our results indicate that TasselNetV4 emerges to be a vision foundation model model for cross-scene, cross-scale, and cross-species plant counting.
To facilitate future plant counting research, we plan to release all the data, annotations, code, and pretrained models at 
\url{https://github.com/tiny-smart/tasselnetv4}.

\end{abstract}

\end{frontmatter}

\section{Introduction}\label{sec:intro}

Plant counting facilitates various tasks in digital agriculture such as plant phenotyping~\citep{khoroshevsky2021parts}, crop growth monitoring~\citep{valente2020automated}, yield estimation~\citep{kitano2019corn}, and crop management~\citep{chamara2023aicropcam}. 
Efficient plant counting can substantially accelerate the cycle of scientific discovery in plant science. For example, counting tiller numbers of rice enables the identification of high-yield varieties, potentially reducing the breeding cycle by up to two years~\citep{lin2020heterosis}. 
Besides scientific research, plant counting is also required in day-to-day agricultural management, where it can face a wide spectrum of spatial scales and imaging conditions, ranging from remote sensing imagery via satellites or unmanned aerial vehicles (UAVs)~\citep{bai2023rice}, to field-based imagery by handheld devices~\citep{buzzy2020real}, and even microscopic imagery in laboratory environments~\citep{zhu2021deep}. 
To meet the need of monitoring large-scale plant populations in diverse scenarios, a number of high-throughput vision-based plant counting approaches have been developed to count plants, such as tomatoes~\citep{rahnemoonfar2017deep}, maize tassels~\citep{hasan2018detection}, and sorghum heads~\citep{ghosal2019weakly}.

\begin{figure}[!t]
    \centering
    \includegraphics[width=\linewidth]{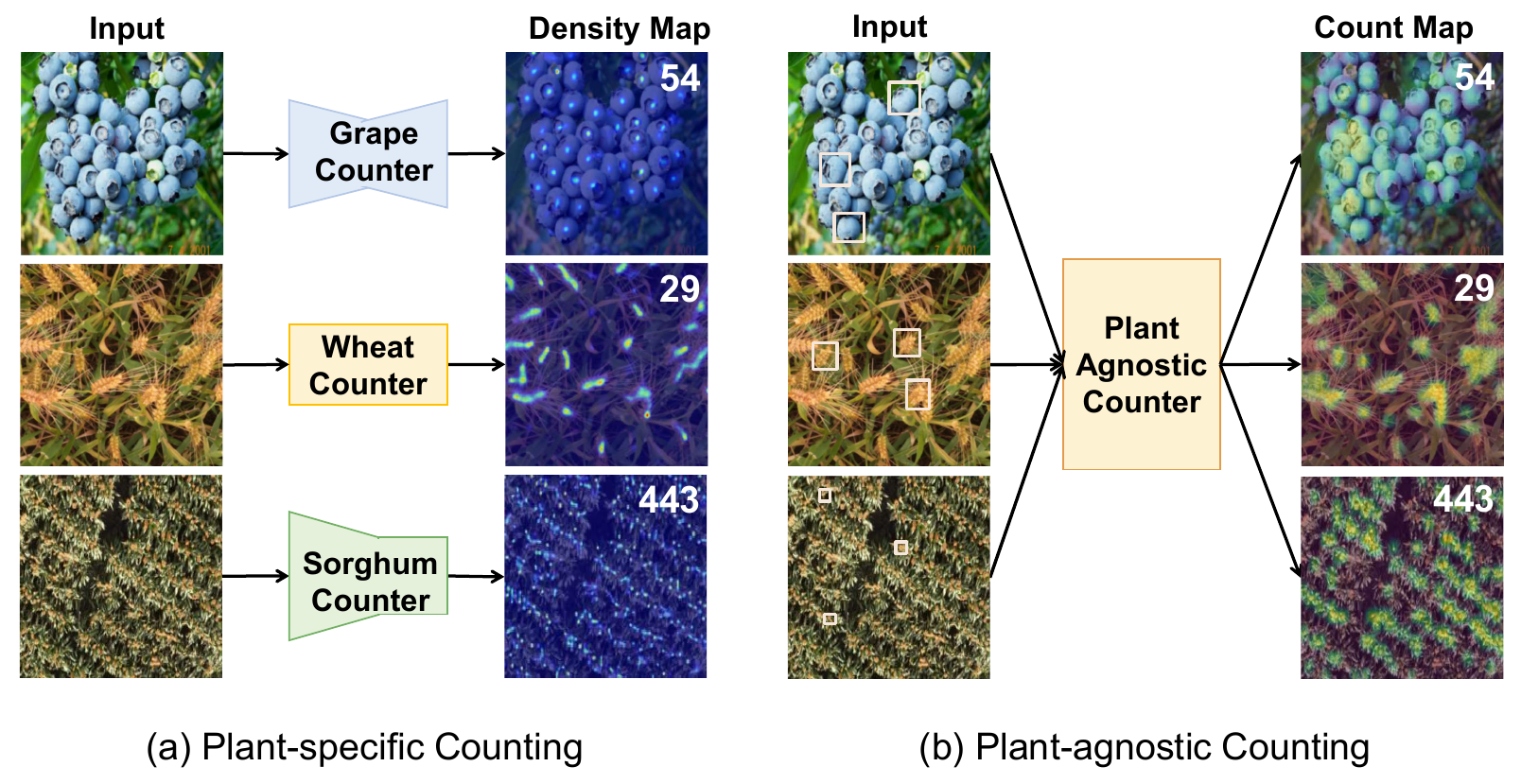}
    \caption{\textbf{Comparison between plant-specific counting and plant-agnostic counting}. In (a) plant-specific counting, each plant species requires training a specific model. For (b) plant-agnostic counting, one could resort to a single model to count any plant species given few plant exemplars (white boxes), without model retraining.}
    \label{fig:intro}
\end{figure}

Albeit effective, existing plant counting models can only count specific plants or organs, such as wheat spikes~\citep{hasan2018detection}, rice plants~\citep{deng2021automated}, and sorghum heads~\citep{ghosal2019weakly}. 
When a target plant changes, the model becomes inapplicable and requires retraining. This implies a repetitive loop of new data collection, data annotation, and model finetuning, introducing excessive costs for practical applications. Indeed a standardizable plant counting model that enables cross-species plant counting is of urgent need for modern agriculture. Inspired by Class-Agnostic Counting (CAC)~\citep{lu2018class} in computer vision, we adapt this concept to the plant domain and introduce Plant-Agnostic Counting (PAC). 
As shown in Fig.~\ref{fig:intro}, in contrast to plant-specific counting that trains separate models for differing plants, PAC aims to count all plants with a single, unified model. 

\begin{figure}[!t]
    \centering
    \includegraphics[width=1.0\linewidth]{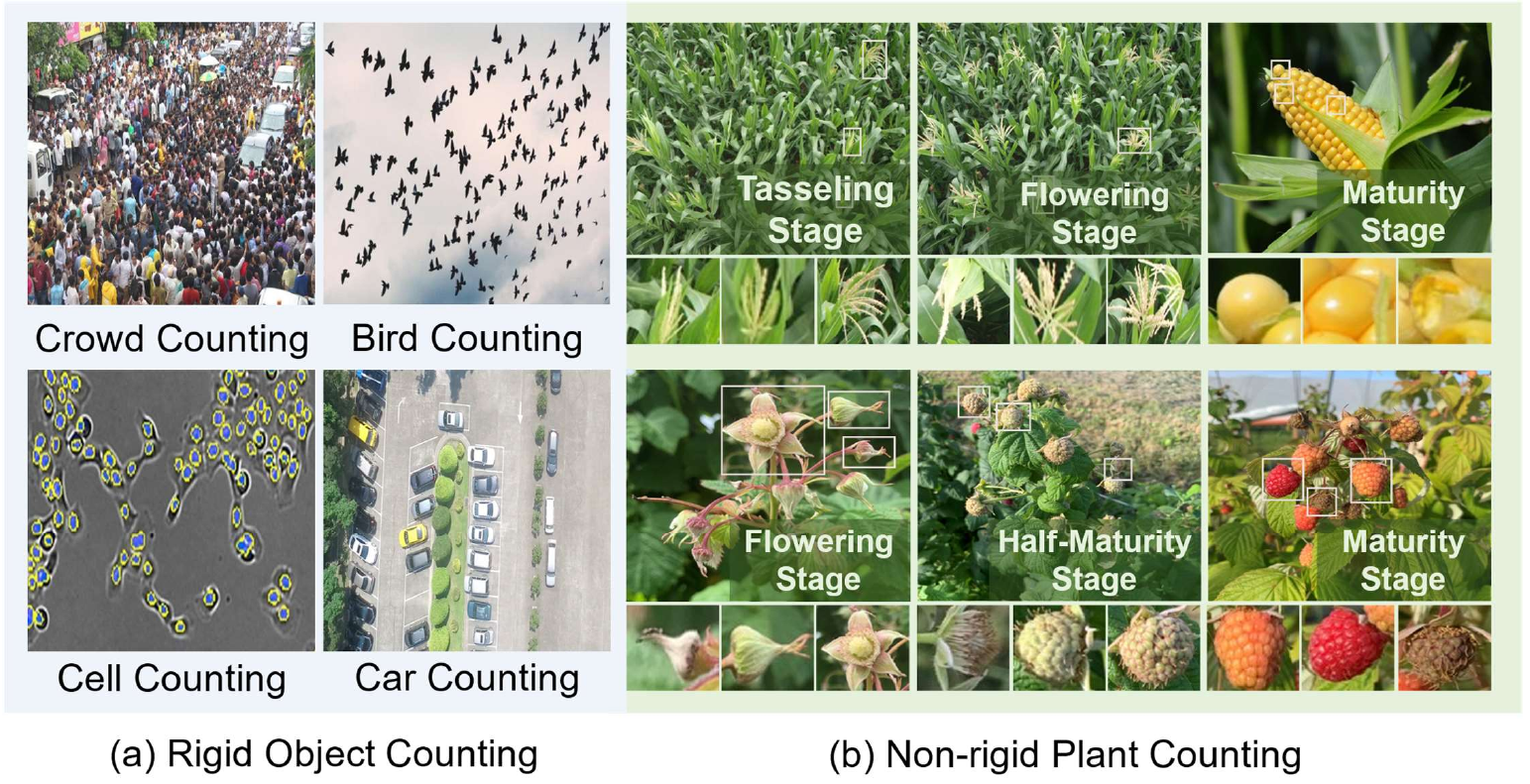}
    \caption{\textbf{Comparison between counting rigid objects and non-rigid plants.} For rigid objects such as crowds, cars, birds, and cells, their shapes and physical sizes do not vary significantly across instances or change greatly with time, while for non-rigid plants, different instances from the same species could have rather different shapes and appearance at different growing stages. The counting instance of interest may also change with time.}
    \label{fig:intro problem}
\end{figure}

Notice that the standard CAC benchmark FSC-147~\citep{famnet} has already included some plant categories\footnote{To clarify the term used, we use \textit{plant species} to indicate different plant varieties or cultivars and \textit{plant categories} to denote different countable categories such as distinct plant organs of the same species.} such as apple and strawberry. Why create another PAC task specific to plants? In this paper, we show that directly adopting existing CAC approaches to PAC can be suboptimal, and vice versa. Current CAC approaches typically focus on measuring the similarity between exemplars and target objects, with an underlying assumption that intra-class variations are much smaller than inter-class ones. 
This assumption may hold for common rigid objects (Fig.~\ref{fig:intro problem}(a)), such as crowds~\citep{zhang2015cross,liu2023point}, cars~\citep{mundhenk2016large,guo2022density}, birds~\citep{akccay2020automated}, and cells~\citep{xue2016cell}, but not for plants. 
Plants are dynamic systems and often present \textit{non-rigid} morphology: their shapes and appearance can change with space (different environmental conditions) and time (different growth stages). The category of interest for counting can also change as plants grow (Fig.~\ref{fig:intro problem}(b)). 
For example, maize tassels are counted during the tasseling and flowering stages, while maize ear kernels are counted at the maturity stage. 
A single plant may have multiple countable instances at different growth stages. They are more than the fruit category covered by FSC-147.

To address the problems above, we draw inspiration from plant-specific counting approaches.
After realizing that the shape and size variations of different plants can cause ambiguous representation in the per-pixel density map (a standard regression target used by mainstream counting approaches), previous plant counting approaches have developed a robust counting paradigm---local counting~\citep{lu2017tasselnet}---to regress the counts directly from sub-image patches. Local counting 
avoids the hard per-pixel learning and alleviates potential occlusions. 
In our previous work, TasselNet~\citep{lu2017tasselnet}, we show that the local counting paradigm works better than the density map regression~\cite{NIPS2010_fe73f687} for plants. 
TasselNetV2~\citep{xiong2019tasselnetv2,lu2020tasselnetv2+} and TasselNetV3~\citep{lu2021tasselnetv3} further advance this paradigm. Given the success of these practices, we presume the local counting idea should also be applicable to PAC.

In this work, we introduce \OurMethod, the new generation of the TasselNet series models, for cross-scene, cross-scale, and cross-species plant counting. Specifically, \OurMethod integrates the local counting idea into the CACViT framework~\citep{wang2024vision}, a clean baseline implemented with a plain vision transformer (ViT), the decoupled attention for implicit similarity matching, and an aspect-ratio scale embedding. Due to the adoption of token-level local counters, we evade the time-consuming ViT decoder blocks and upsampling stages in CACViT. 
To enhance cross-scale generalization, we propose multi-branch box-aware counters that dynamically adapt to different plant scales conditioned on exemplar sizes. 
We remark that, different from the physical-size variations derived from the non-rigid nature of plants, plants can suffer from \textit{scale} variations as well due to the perspective change and varying imaging heights. 
To make full use of the attention map, we also develop an attention-informed visualizer to visualize counting results. 
To facilitate the study of PAC and to validate our approach, we construct the first two challenging plant-agnostic counting datasets, PAC-$105$ and PAC-Somalia. 
PAC-$105$ encompasses $2,646$ images and $105$ countable plant categories from $64$ plant species, including fruits, seeds, flowers, and leaves, and additionally, covers morphologically diverse plants across various growth stages. 
PAC-Somalia comprises $1,542$ images from $32$ exclusive plant species in Somalia. Most of these images are not available online and are all manually captured due to the difficulty of data collection, which is thus particularly valuable and suitable for out-of-distribution (OOD) evaluation.

Extensive experiments under both the $1$-shot and $3$-shot settings demonstrate that \OurMethod achieves state-of-the-art performance on the PAC-$105$ and PAC-Somalia datasets. Under the $3$-shot setting, \OurMethod achieves an MAE (Mean Absolute Error) of $16.04$, an RMSE (Root Mean Square Error) of $28.03$, a WCA (Weighted Counting Accuracy) of $0.74$, and an $R^2$ (R-squared) value of $0.92$ on the PAC-$105$ dataset. Additional results on the PAC-Somalia dataset also confirm the superiority of our approach, achieving an MAE of $8.88$, an RMSE of $13.11$, a WCA of $0.72$, and an $R^2$ value of $0.87$. Visualization results indicate that our model is capable of counting a target plant from complex background and of discriminating a target plant from multiple plant species. Moreover, the model exhibits strong robustness to scale variations, producing accurate predictions across varying scales, including both remote sensing and in-field imagery.
In terms of efficiency, our method also achieves a runtime reduction of approximately $35.67\%$ and reduces parameter counts by $18.10\%$ compared with the standard CACViT baseline. 

The contributions of this work include the following: 
\begin{itemize}[leftmargin=*]
    \item PAC: a new plant-orientated task called plant-agnostic counting, customizing class-agnostic counting into the plant domain and highlighting zero-shot generalization across fine-grained plant species; 
    \item TasselNetV4: a vision foundation model that inherits the vein of TasselNet plant counting model series and that extends plant-specific counting to cross-scene, cross-scale, and cross-species plant counting; 
    \item PAC-$105$ and PAC-Somalia: two challenging PAC datasets, which provides comprehensive benchmarks and facilitates future plant counting studies.

\end{itemize}

\section{Related work}\label{sec:relate work}

\subsection{Plant counting}\label{sec:r.plant counting}

The applications of computer vision in plant counting can date back to 2016~\citep{giuffrida2016learning}, where a learning-based model is proposed to count the leaves of rosette plants. Simultaneously, DeepCount~\citep{rahnemoonfar2017deep} applies a deep learning model to count tomatoes. Both methods, however, were tested in controlled indoor environments. DeepFruit~\citep{bargoti2017deep} marked the first attempt to count plants in the field, using a CNN-based method to detect and sum individual tomatoes in an orchard.

TasselNet~\citep{lu2017tasselnet}, a milestone in plant counting, showcases one of the first applications of modern deep regression techniques in plant counting. It also triggers the debate of two mainstream plant counting paradigms: detection-based counting and regression-based counting.
Several approaches have been developed following the two paradigms, managing to count leaves~\citep{itzhaky2018leaf}, sorghum heads~\citep{ghosal2019weakly}, wheat spikes ~\citep{hasan2018detection}, and so on. TasselNetV2~\citep{xiong2019tasselnetv2} and TasselNetV2+~\citep{lu2020tasselnetv2+} emphasize the benefits of regression-based methods with local context, especially in cases with overlaps and varying object sizes typical in plant counting. Despite state-of-the-art efficiency and accuracy, redundancy in count maps hampers visualization clarity. To address this, TasselNetV3~\citep{lu2021tasselnetv3} proposes an encoder-decoder architecture with guided upsampling, leading to improved explainability. Detection-based methods~\citep{yang2021rapid,lu2024maize,li2023tomato,lu2023plant} have gained popularity due to YOLO~\citep{redmon2016you}. While they typically outperform the TasselNet series, the improvement comes at the cost of more expensive bounding box annotations~\citep{xiong2019tasselnetv2}, compared to dotted annotations used by regression-based approaches. 

Looking back to these methods, they generally follow a similar workflow: targeting a specific plant, collecting new data of the plant, annotating the data, and training a new model. Such a plant-specific workflow leads to the repetitive execution of time-consuming tasks, and limits the standardization of a model to promote to wide plant species.

\subsection{Class-agnostic counting}\label{sec:r.CAC}

\begin{figure*}[!t]
    \centering
    \includegraphics[width=\textwidth]
    {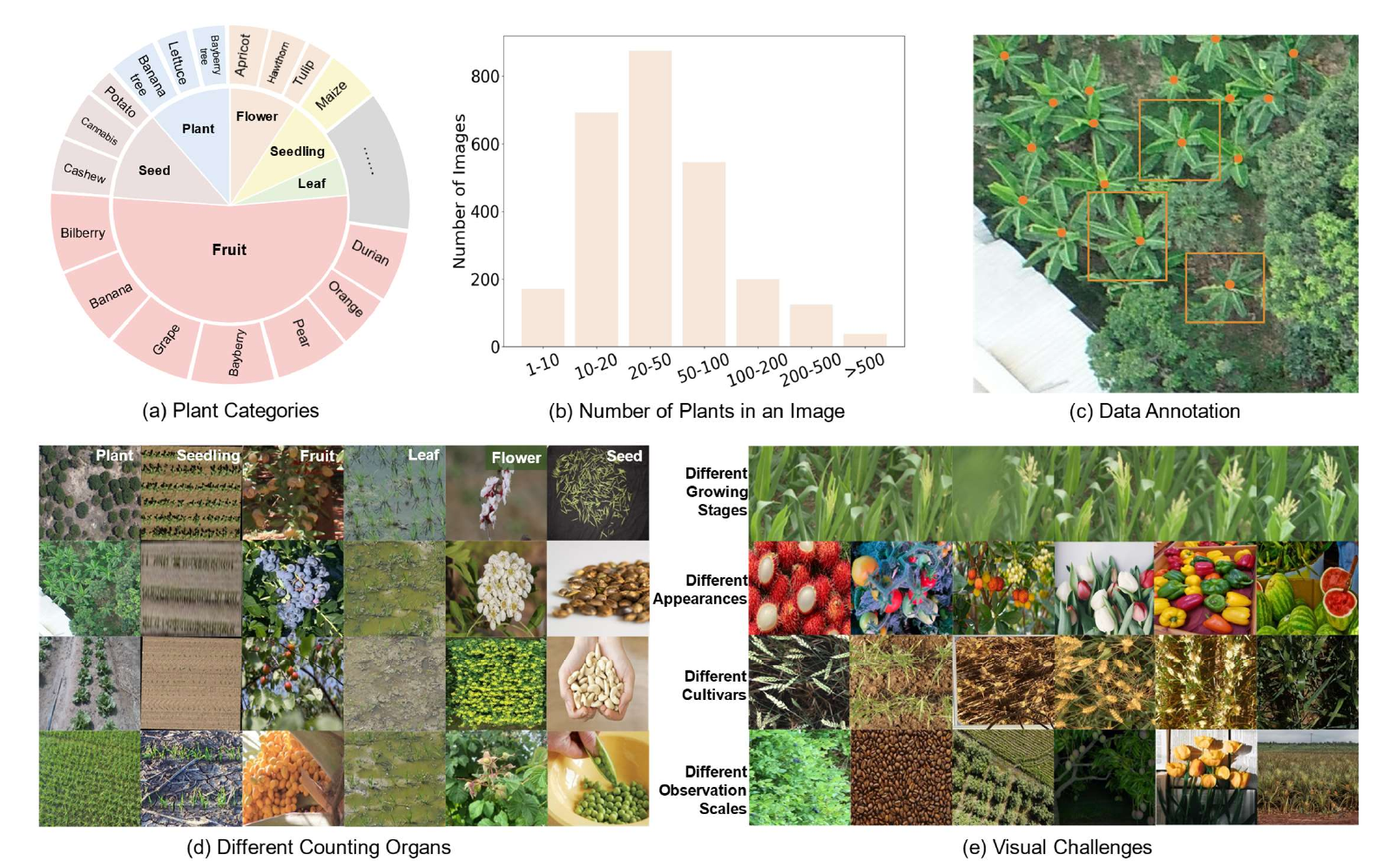}
    \caption{\textbf{Overview of the PAC-$105$ dataset.} (a) Distribution of different plant categories. (b) Number of plants in an image. (c) Data annotation process. Each instance is labeled with a dot, while three instances are boxed as exemplars. (d) Example images in the dataset. (e) Visual challenges in the dataset. First three rows show instances with large intra-class variations, such as different stages (row 1), different colors (row 2), and different cultivars of a single species (row 3). The last row shows different imaging conditions, including extremely sparse and dense scenes (column 1 and 2), remote sensing imagery (column 3), poor lighting (column 4), indoor environment (column 5) and outdoor environment (column 6).}
    \label{fig:mm.dataset}
    
\end{figure*}
\begin{figure}[!t]
    \centering
    \includegraphics[width=\linewidth]
    {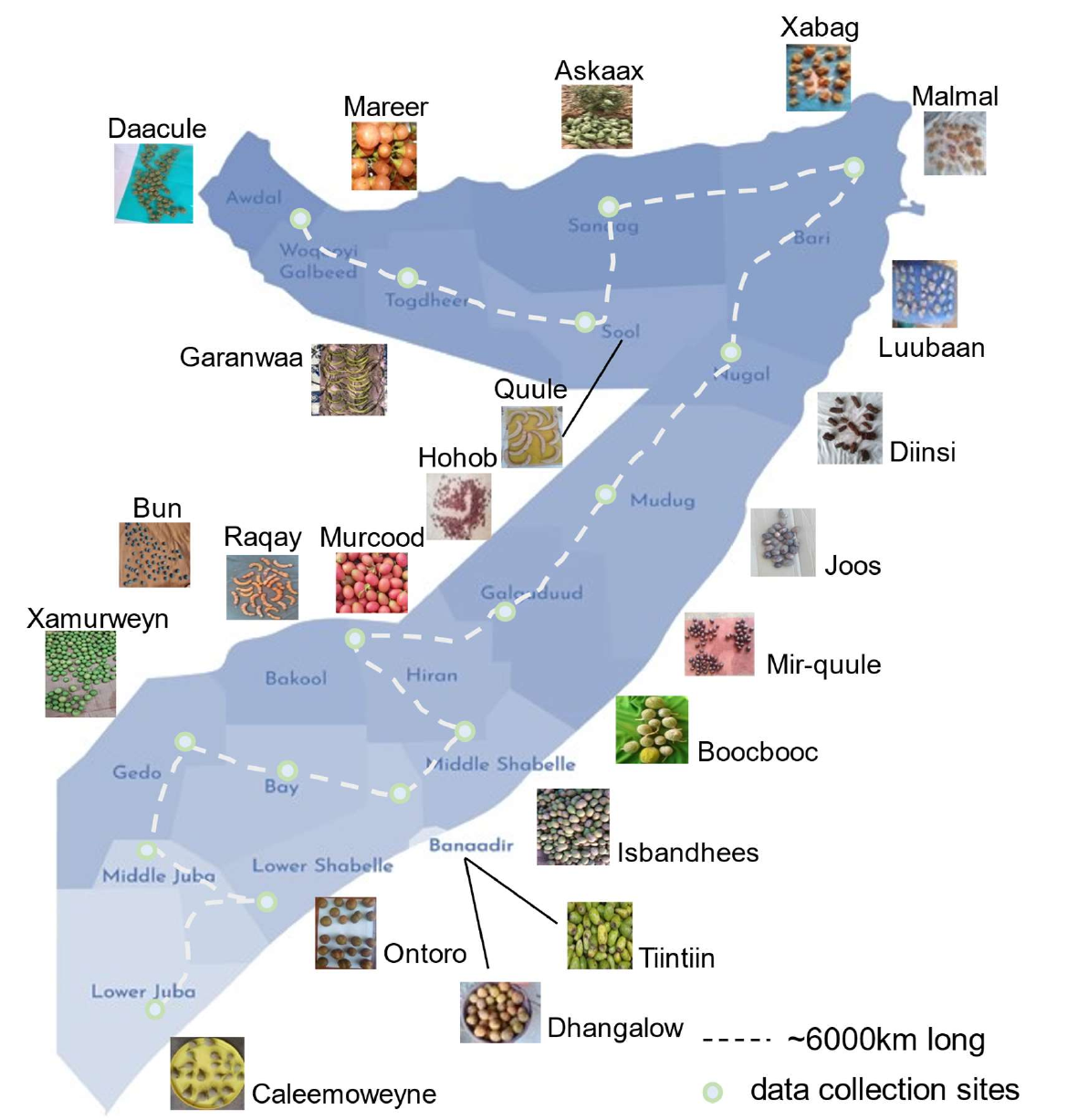}
    \caption{\textbf{Unique species from Somalia in the PAC-Somalia dataset.} Local names are given for species without scientific names. }
    \label{fig:mm.dataset-somalia}
\end{figure}

Counting objects from an image has been a long-standing problem in computer vision. 
Existing approaches mainly focus on counting specific object classes~\citep{zhang2015cross,guo2022density,xue2016cell}. To mitigate class dependency, a new direction of CAC emerges recently~\citep{lu2018class}, aiming to build a counting model that generalizes to wide object categories. A dataset called FSC-$147$ is also constructed for CAC. CAC is typically formulated in an `extract-then-match' manner. \cite{lu2018class} applies a CNN to extract both image and exemplar features and uses inner product to match them. \cite{radford2021learning} further introduces a correlation map to enable multi-scale matching. \cite{shi2022represent} introduces similarity guidance from a single matching stage to two stages, with a dynamic similarity metric and a similarity loss. Instead of using the similarity guidance implicitly, \cite{you2023few} explicitly enhance the regression features, and \cite{lin2022scale} introduces a scale prior to the deformable convolution block. 

With Vision Transformer (ViT,~\cite{dosovitskiy2020image}) becoming off-the-shelf, there is a growing interest in integrating the attention mechanism in CAC. \cite{liu2022countr} first uses self-attention blocks for feature extraction and then treats images as queries and exemplars as keys and values to execute cross-attention for matching. \cite{wang2024satcount} embeds size information into ViT, helping respond to the scale variations of different instances. \cite{wang2024vision} instead argues that it is more straightforward for ViT to model images and exemplars in the same token sequence rather than separately as in CountTR. They expound this idea as `extract and match' with a decoupled view of attention and report state-of-the-art performance on the FSC-$147$ dataset.

In this work, we introduce the PAC task which not only extends the problem connotation of CAC but also imposes new challenges outside the standard CAC.

\section{Materials and methods}\label{sec:M&M}
In this section, we first introduce two new PAC datasets, PAC-$105$ and PAC-Somalia, constructed following the principles of biodiversity and environmental diversity. We then formulate the class-agnostic counting problem, introduce our baseline model CACViT, and elaborate on our improvements towards PAC in detail. 

\subsection{PAC-$105$ and PAC-Somalia datasets}
In this work, we develope two datasets, including the PAC-$105$ dataset and the PAC-Somalia dataset.
They are built with the following principles in mind: 

\begin{itemize}
    \item \textbf{Diversity between species.} The dataset should cover a sufficient number of plant species to reflect biodiversity;
    \item \textbf{Diversity within species.} For a specific plant, the samples should cover sufficient differences in appearance, growth stages, and other intrinsic factors;
    \item \textbf{Diversity across scenes.} The dataset should cover different real-world environmental variations such as weather and soil conditions, including not only controlled laboratory environments but also field-based ones, with complex backgrounds, differing illumination, various growth stages, and so on;
    \item \textbf{Diversity across scales.} The dataset should feature diverse observation scales, encompassing hand-held imagery (smartphones), ground-based imagery (fixed digital cameras), and remote sensing imagery (drones and satellites).  
\end{itemize}
Following these principles, we construct two datasets, with a broad range of species, visual appearance, and environmental variations. We reveal the data collection and annotation details as follows.

\paragraph{PAC-$105$ dataset}
The PAC-$105$ dataset consists of $2,646$ images and features $64$ plant species with $105$ different countable cultivars, as shown in Fig.~\ref{fig:mm.dataset}. 
The images are classified based on the target organ and plant species.
Each countable category contains a minimum of $4$ and a maximum of $50$ images, covering a wide range of cereal crops (e.g., wheat and rice), ornamental plants (e.g., tulip), vegetables (e.g., lettuce), fruits (e.g., apple and orange), industrial plants (e.g., beet and rapeseed), and different organs of the plants, including the whole plant, seedling, fruit, leaf tip, seed, flower, and so on. 
The image data are collected from multiple sources to guarantee data diversity, including existing publicly available datasets for plant detection~\citep{zou2020maize, david2020global}, plant counting~\citep{lu2017tasselnet,lu2021tasselnetv3, guo2018aerial}, plant classification datasets~\citep{lin2014microsoft}, and open-sourced images from the Internet.
The dataset is split into a training set and a testing set with \textit{disjoint} plant-level or organ-level categories.   
In particular, $65$ categories are assigned to the training set, and the remaining $40$ categories are assigned to the testing set, with $1,656$ and $991$ images, respectively.

We highlight the distinguished high intra-class variations in our dataset. As depicted in Fig.~\ref{fig:mm.dataset}(e), this intra-class variation results mainly from three aspects: different growth stages, phenotypic variations among different plant individuals of the same species (e.g., color and shape) and of different cultivars, and diversity in scenarios, including different weather conditions (sunny, cloudy, rainy, and overcast), observation scales (remote sensing and ground-based measurements), and plant distributions (sparse and dense).

We annotate our dataset following the FSC-$147$ dataset and we use the VGG Image Annotator~\citep{VGG2023VIA} for annotation. Each instance is marked with a dot, and three exemplars are given with bounding boxes. 
Specifically, dot annotations are marked around the center of each instance. For occlusion cases, the occluded instance is annotated only when the occlusion is less than $90\%$. Three exemplars are chosen randomly. 

\paragraph{PAC-Somalia dataset} 
We also collect a nationwide PAC-Somalia dataset for OOD evaluation. This dataset includes $1,542$ images from $32$ unique plant species in Somalia. To maximize species diversity, we include at least one species unique to the local ecosystem in each sub-region of Somalia, as shown in Fig.~\ref{fig:mm.dataset-somalia}. All images are manually captured, with around $6000$ kilometers traveling throughout the country, including also multiple growth stages, a variety of viewing angles, heterogeneous lighting conditions, and differing spatial densities. 
We annotate this dataset following the same principles as PAC-$105$. 
Notably, plant species in PAC-Somalia are underrepresented in public data, including PAC-$105$, such that they are neither visible to the training set nor to pretrained backbones, 
making the dataset suitable for OOD evaluation. 

\subsection{Problem formulation of few-shot class-agnostic counting}
Class-agnostic counting (CAC) aims to count objects of interest in an image without predefined categories~\citep{lu2018class}. In practice, the object of interest is specified by a user-provided bounding box from the input image, a.k.a. the exemplar. Formally, given an input image $\bm{I} \in \mathbb{R}^{3 \times H \times W}$ of height $H$ and width $W$ and $l$ exemplars $\bm{I}_e \in \mathbb{R}^{l\times 3 \times h \times w}$ of height $h$ and of width $w$ (exemplars are resized here), indicating the object of interest, the goal of CAC is to learn a function $\mathcal{F}_\theta$ parameterized by $\theta$ that satisfies  
\begin{equation}
    \label{equ:mm.cac}
    \bm C=\mathcal{F}_\theta(\bm{I},\bm{I}_e)\,,
\end{equation}
where $\bm C$ can be any form of counting-related output such as a density map~\citep{famnet}, a set of bounding boxes~\citep{jiang2024t}, or a count map~\citep{lu2017tasselnet} used by this work.

\subsection{CACViT baseline}\label{sec:mm.CACVtT}

\OurMethod builds on our previous work CACViT~\citep{wang2024vision}, a state-of-the-art class-agnostic counter based on a plain vision transformer.
CACViT takes the image $\bm{I}$, the exemplars $\bm{I}_e$, and an additional box-informed scale embedding $\bm{S}$ as input, regressing the density map to represent object counts.  

\paragraph{Decoupled attention}
At the core of CACViT is the proposition of decoupled attention. To exposit it, we begin with the self-attention~\citep{vaswani2017attention} in vision transformer.  
Assume that an image $\bm{I}$ is tokenized to $N$ tokens by a patch size $p$ and generate a triplet of $d$-dimensional query $\bm{Q}$, key $\bm{K}$, and value $\bm{V}$. They form the self-attention mechanism
\begin{equation}\label{equ:mm.amap}
    \text{Attention}(\bm{Q},\bm{K},\bm{V})=\text{softmax}\left(\frac{\bm{A}(\bm{Q},\bm{K})}{\sqrt{d}}\right) \bm{V}\,,
\end{equation}
where $N=\frac{HW}{p^2}$, and $\bm{A}(\bm{Q},\bm{K})=\bm{Q} \bm{K}^T$ denotes the attention map of size $N \times N$. Given Eq.~\eqref{equ:mm.amap}, the attention-weighted value vector $\bm{v}_i^{\prime}$ can be computed by 
\begin{equation} \label{equ:mm.vector atten}
    \bm{v}_i^{\prime}= \frac{\sum_{j=1}^{N}sim(\bm{q}_i,\bm{k}_j)~\bm{v}_j}{{\sum_{j=1}^{N}sim(\bm{q}_i, \bm{k}_j)}}\,,
\end{equation}
where $\bm{q}_i$, $\bm{k}_i$, and $\bm{v}_j$ correspond to the $i$-th query vector, the $k$-th key vector, and the $j$-th value vector, respectively, and $sim(\bm q,\bm k)=\exp(\frac{\bm q^T\bm k}{\sqrt{d}})$ computes the similarity between query and key vectors. Eq.~\eqref{equ:mm.vector atten} unifies self-attention and cross-attention. When $\bm{q}_i$, $\bm{k}_i$, and $\bm{v}_j$ are from the same token sequence, it is self-attention; when $\bm{q}_i$ is from another query token sequence, it becomes cross-attention. 

\begin{figure}[!t]
    \centering
    \includegraphics[width=1.0\linewidth]{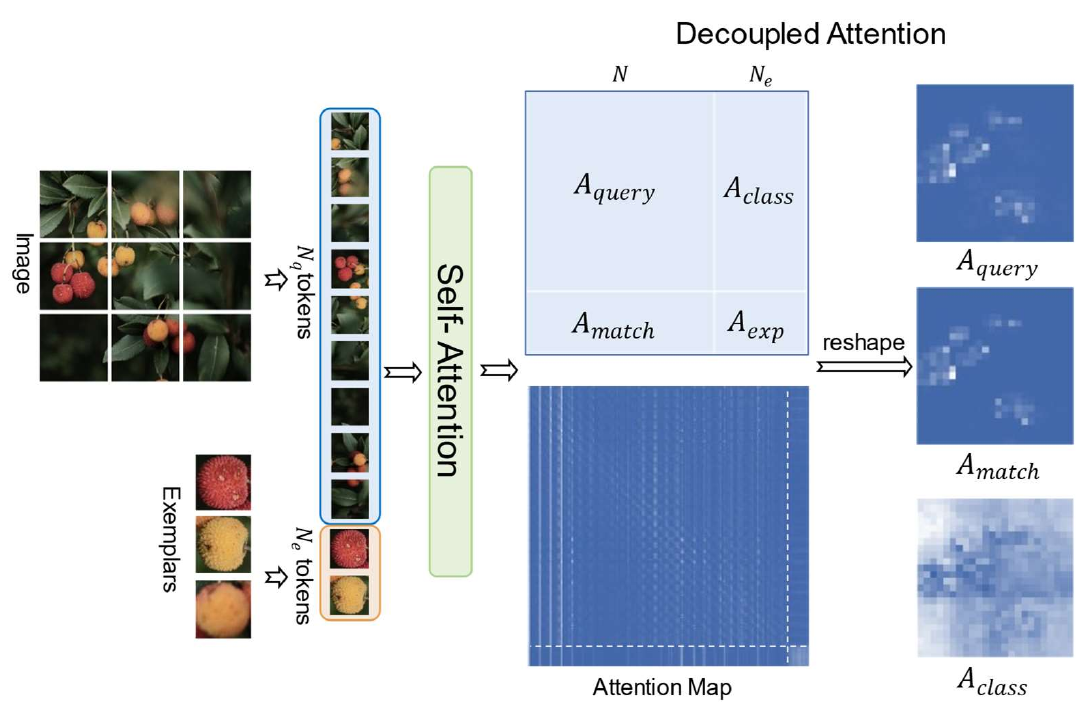}
    \caption{\textbf{Decoupling, generation, and visualization of attention map in CACViT.} $\bm{A}_{match}$ and $\bm{A}_{query}$ highlight the foreground, while $\bm{A}_{class}$ highlights the background. } 
    \label{fig:mm.attention_decouple}
\end{figure}

Given image and exemplar tokens, one can execute unified self-attention and cross-attention using Eq.~\eqref{equ:mm.vector atten}. For a sequence of $N=N_q + N_e$ tokens, where $N_q$ and $N_e$ represent the number of image and exemplar tokens respectively, attention at this level, shown in Fig.~\ref{fig:mm.attention_decouple}, functions as both self-attention and cross-attention from a \textit{decoupled} perspective as
\begin{equation}\label{equ:mm.decouple attention}
    \bm{A} = \begin{bmatrix}
        \bm{A}_{query} && \bm{A}_{class} \\
        \bm{A}_{match} && \bm{A}_{exp}
    \end{bmatrix}\,,
\end{equation}
where $\bm{A}_{query}\in \mathbb{R}^{N_q\times N_q}$ and $\bm{A}_{exp} \in \mathbb{R}^{N_e \times N_e}$ denote the self-attention map w.r.t. the image and the exemplar, respectively, $\bm{A}_{match} \in \mathbb{R}^{N_e \times N_q}$ denotes the image-to-exemplar cross-attention map, and $\bm{A}_{class} \in \mathbb{R}^{N_q \times N_e}$ denotes the exemplar-to-image cross-attention map.
This decoupled view achieves joint feature enhancement and condition injection, allowing implicit matching between image and exemplar tokens. 
This `extract-and-match' operation has proven to be more effective than the previous `extract-then-match' paradigm~\citep{famnet,shi2022represent}. 

\paragraph{Aspect-ratio-aware scale embedding}

To compensate for the loss of both scale and magnitude, CACViT introduces an aspect-ratio-aware scale embedding approach for each exemplar. Given $l$ exemplars, the $i$-th exemplar is of size $3 \times h_i \times w_i$, and it is resized to a fixed size of $3 \times h \times w$ before tokenization, leading to the information loss of both aspect ratio and scale. 
CACViT generates a width map $\bm{S}_i^{w} \in \mathbb{R}^{h \times w}$ with column-vector values ranging from $0$ to $w_i$ and a height map $\bm{S}_i^{h} \in \mathbb{R}^{h \times w}$ with row-vector values ranging from $0$ to $h_i$, forming the scale embedding $\bm{S}_{i} \in \mathbb{R}^{h\times w}$ to inform the scale and aspect ratio, formulated by
\begin{equation}\label{eq:scale-embedding}
    \bm S_{i} = \bm{S}_i^{h} + \bm{S}_i^{w}\,,
\end{equation}
where $\bm{S}^{w}_{i}(m,n) = n\frac{w}{w_i}$ and $\bm{S}^{h}_{i}(m,n) = m\frac{h}{h_i}$. All resized exemplars and their corresponding scale embeddings are concatenated as input to CACViT.

To compensate for the magnitude loss, the size information is further passed to the ViT block to enhance the attention map. Given resized exemplars $\bm{I}_{e}\in \mathbb{R}^{l\times 3 \times h\times w}$, we represent the magnitude embedding $M_e$ as:
\begin{equation}\label{equ:magnitude}
    M_e = \frac{1}{l}\sum_{i=1}^{l}\frac{w h}{w_i  h_i}\,.
\end{equation}
This variable serves as a scaling factor of $\bm{A}_{match}$ to compensate for the magnitude loss caused by softmax.

\begin{figure*}[!t]
    \centering
    \includegraphics[width=1.0\textwidth]{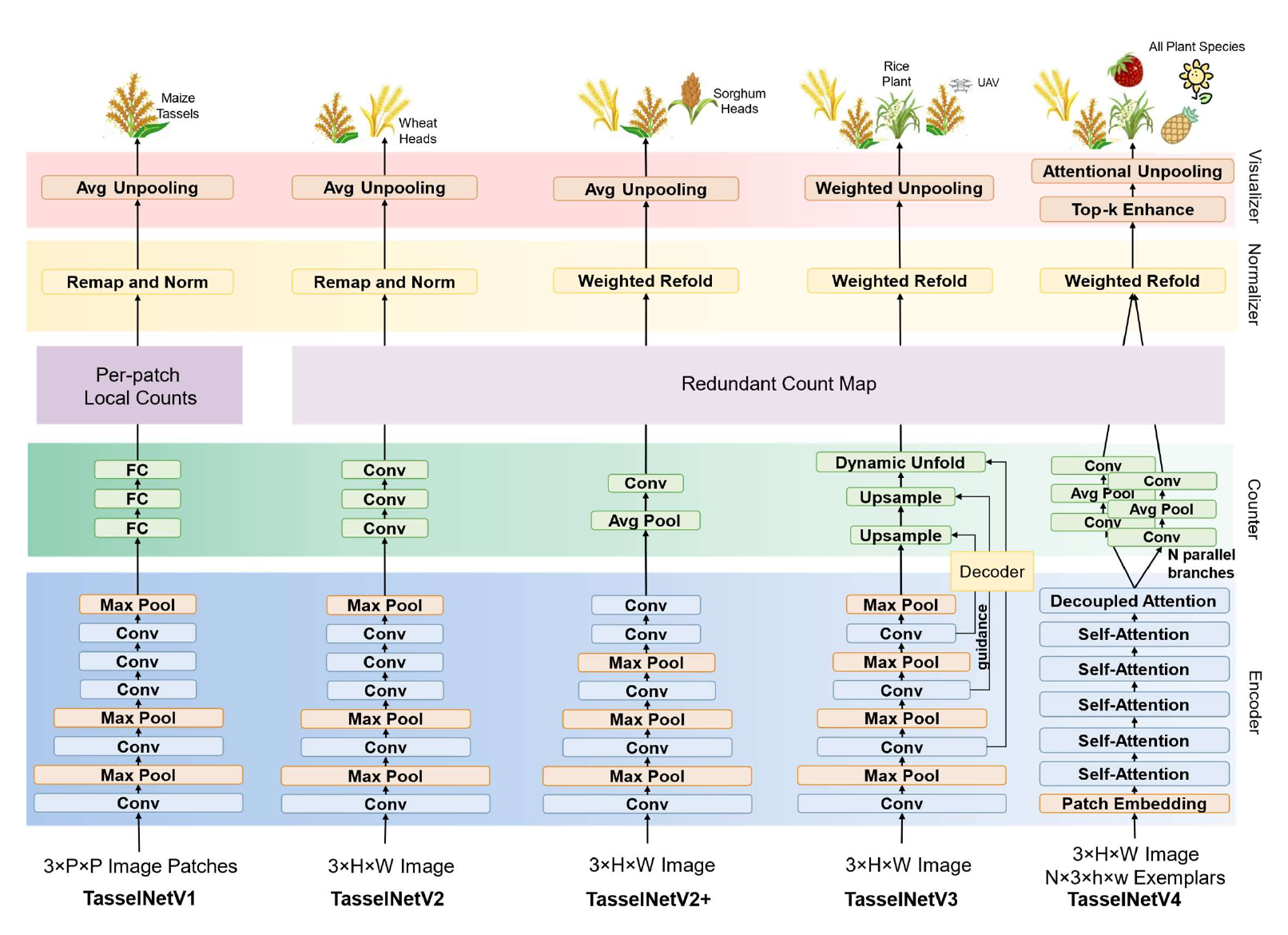}
    \caption{\textbf{Comparison between TasselNets.} \textit{TasselNetV1} utilizes a CNN network to regress per-patch local counts and obtains the count map by spatial concatenating after prediction, successfully applied to maize tassel counting. \textit{TasselNetV2} presents a fully convolutional local counter to process the whole input image, extending its applicability to wheat head counting. \textit{TasselNetV2+} proposes a GPU-based normalizer to speed up inference and adopts a new encoder and counter. It manages to count sorghum heads additionally. \textit{TasselNetV3} constructs an encoder-decoder framework to enhance the explainability of count map visualization. The new designed architecture expands its usage to more plants, including rice plant counting. \textit{\OurMethod} integrates local count with ViT framework, managing to count plant regardless of species.}
    \label{fig:mm.Tasselnet}
\end{figure*}
\paragraph{Deficiency of unreasonable tokenization}
While benefiting from the simplicity of ViT, 
CACViT inherits a limitation of large-stride patch tokenization ($\frac{1}{16}$ resolution of the original image). The fixed stride and limited resolution can cause problems when using a local counter and tackling plants of multiple scales. 
Next we elaborate on how we address these problems with multi-branch box-aware token-level counters.

\subsection{\OurMethod for plant-agnostic counting}\label{sec:TN4}
Before presenting \OurMethod, we first briefly recap the trajectory from TasselNet to TasselNetV3.

\subsubsection{From TasselNet to TasselNetV3}
TasselNet is a specialized framework designed for plant counting. It leverages the idea of redundant local counting to address specific challenges in plant counting, such as occlusions, plant growth, and high intra-class variability. 
Basically, this framework comprises several key modules, including an encoder $\mathcal{E}$ used to extract image features, a decoder $\mathcal{D}$ used to recover image resolution and refine features, a counter $\mathcal{C}$ used to output redundant count values, a normalizer $\mathcal{N}$ used for de-redundancy, and a visualizer $\mathcal{V}$ used to visualize counting results. The decoder is optional.
Fig.~\ref{fig:mm.Tasselnet} compares different TasselNets. 

\paragraph{TasselNetV1}

Being one of the first deep regression based plant counting models, TasselNet~\citep{lu2017tasselnet} manages to count maize tassels in the field using a CNN encoder $\mathcal{E}_1$ and a simple local patch counter $\mathcal{C}_1$ implemented by three fully-connected layers, regressing the per-patch count $r$, which takes the form
\begin{equation}
    \bm C = {\mathcal{N}_1}\left({\tt \underset{i}{cat}}~\mathcal{C}_1(\mathcal{E}_1(\bm{I}_i))\right)\,,
\end{equation}
where $\bm{I}_i$ is the $i$-th local image patch, and ${\tt \underset{i}{cat}}$ accumulates and projects all elements back to a map indexed by $i$. The ${\tt cat}$ operator is used to accumulate overlapped image patches and their local counts to form a redundant count map $\bm R$, and the normalizer $\mathcal{N}_1$ averages each per-patch count in $\bm R$ into per-pixel counts divided by the patch counting frequency to generate the normalized count map $\bm C$. For output visualization, a visualizer $\mathcal{V}_1$ applies spatial average unpooling to redistribute each local count to a local region. TasselNetV1 points out a promising direction for addressing non-rigid plant counting over conventional density based approaches. 

\paragraph{TasselNetV2}
To make full use of the local context in TasselNet, 
TasselNetV2~\citep{xiong2019tasselnetv2} 
processes the input image $\bm I$ in a fully convolutional manner and accurately counts wheat spikes and maize tassels. It is formulated by
\begin{equation}
    \bm C = \mathcal{N}_1\left(\mathcal{C}_2\left(\mathcal{E}_1(\bm I)\right)\right)\,,
\end{equation}
where $\mathcal{C}_2$ denotes the TasselNetV2 counter implemented by equivalent $1\times 1$ convolutional layers. For de-redundancy, the same normalizer $\mathcal{N}_1$ is also applied to obtain the count map $\bm C$, followed by the same visualizer $\mathcal{V}_1$.  
TasselNetV2 marks the basic maturity of modular design of the framework, featuring an encoder, a counter, a normalizer, and a visualizer in sequence. 

\begin{figure*}[t]
    \centering
    \includegraphics[width=\textwidth]
    {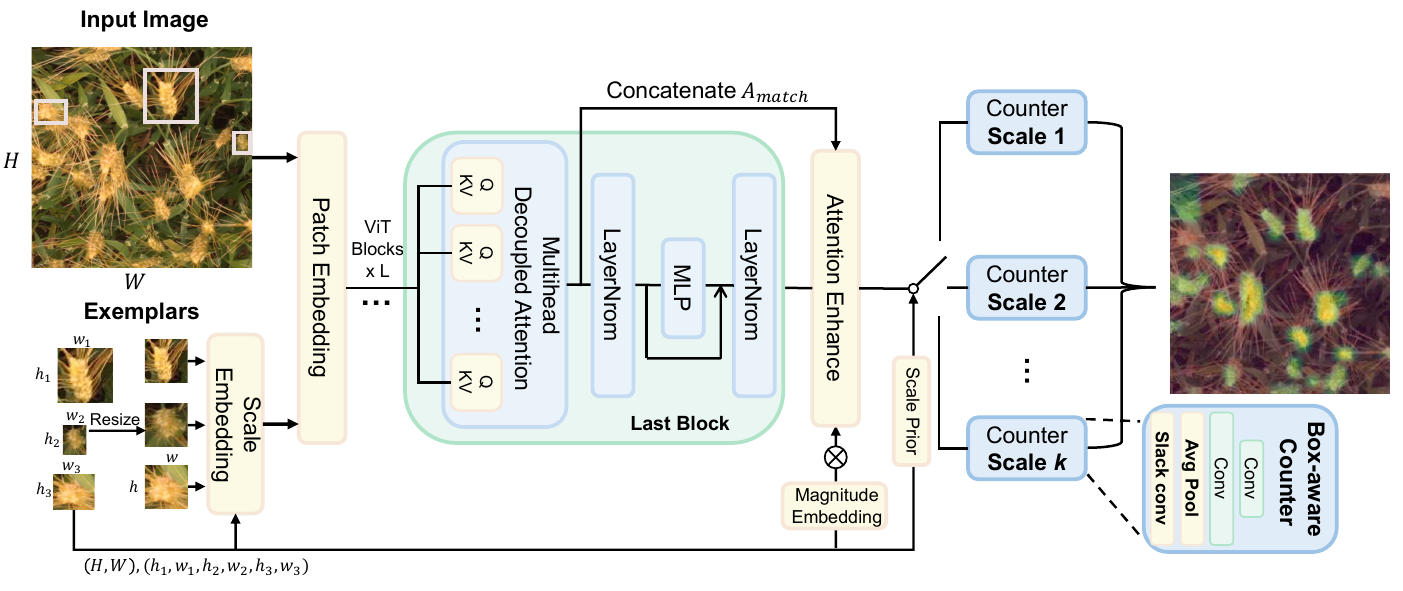}
    \caption{\textbf{Framework of \OurMethod.} An input image and query exemplars are first tokenized into patches to form a joint token sequence. Meanwhile, the width and height of image and exemplars are passed to compute the magnitude embedding (Eq.~\eqref{equ:magnitude}) and scale prior (Eq.~\eqref{equ: scale rule}). After several ViT blocks, the last block decouples the attention to obtain the $A_{match}$ (Fig.~\ref{fig:mm.attention_decouple}). By further enhancing the image features with $A_{match}$ and compensating with the magnitude embedding, the enhanced features are finally decoded with multi-branch local counters. For each image, an appropriate local counter is selected based on the scale prior to generate the final output.}
    \label{fig:mm.TasselNetV4}
\end{figure*}

\paragraph{TasselNetV2+}
To improve efficiency, TasselNetV2+~\citep{lu2020tasselnetv2+} profiles TasselNetV2 and identifies the bottleneck of the normalizer, therefore proposing a GPU-based implementation $\mathcal{N}_2$, reformulating the $\mathcal{N}_1$ with GPU-based operator. It also simplifies the encoder $\mathcal{E}_1$ with a pre-downsampled encoder $\mathcal{E}_2$ and introduces a lightweight local counter $C_3$ with parameter-efficient global average pooling. The model demonstrates lossless performance on counting maize tassels, wheat spikes, and sorghum heads. The whole pipeline takes the form 
\begin{equation}
    \bm C = \mathcal{N}_2(\mathcal{C}_3\left(\mathcal{E}_2(\bm I)\right))\,.
\end{equation}
Despite being much faster, the visualizer $\mathcal{V}_1$ remains unsatisfactory for intuitive count visualization.

\paragraph{TasselNetV3}
To enhance the explainability of the count map, TasselNetV3~\citep{lu2021tasselnetv3} attributes the poor visualization to spatially averaged unpooling, which hinders precise localization.
To address this, TasselNetV3 introduces a decoder $\mathcal{D}_1$ to generate the feature map of low resolution $\bm F_l$ and of high resolution $\bm F_h$, that is,
\begin{equation}
    \bm F_l\, , \bm F_h = \mathcal{D}_1(\mathcal{E}_3(\bm I))\,.
\end{equation}
Then TasselNetV3 proposes a novel counter $\mathcal C_4$ that leverages $\bm F_h$ to guide the re-distribution of local counts dynamically, which amounts to
\begin{equation}
    \bm C = \mathcal{N}_3\left(\mathcal{C}_4(\bm F_l, \bm F_h)\right)\,.
\end{equation}
With the new counter, TasselNetV3 introduces a new visualizer $\mathcal{V}_2$ that adopts $F_h$ to improve the fidelity of the count map, thus improving the explanability. It also expands the application scenarios to counting rice plants and counting maize tassels captured by UAVs. 

\paragraph{Remark} Although these TasselNet series models have shown promising results, they always need to train a new model for a new plant species. We address this dilemma in a novel plant counting framework \OurMethod.

\subsubsection{TasselNetV4 architecture}

TasselNetV4 follows the spirit of TasselNet and adopts a similar framework. Yet, to enable plant-agnostic counting, it significantly interacts the CACViT framework. Concretely, TasselNetV4 consists of a plain vision transformer as the encoder $\mathcal{E}_4$, a novel token-level multi-branch box-aware local counter $\mathcal{C}_5$, a conventional normalizer $\mathcal{N}_2$, and an attention-informed visualizer $\mathcal{V}_3$. Fig.~\ref{fig:mm.TasselNetV4} illustrates the TasselNetV4 pipeline, and Fig.~\ref{fig:mm.Tasselnet} compares it with previous versions.

\paragraph{Network overview}
 
Akin to CACViT, TasselNetV4 considers a query image $\bm{I}$, the exemplars $\bm{I}_e$, and the scale embedding $\bm{S}$ as the inputs. Yet, different from CACViT, the goal turns to regress the normalized count map, formulated by
\begin{equation}
    \bm C = \mathcal{N}_2(\mathcal{C}_5(\mathcal{E}_4(\bm I, \bm{I}_e,\bm S),\bm S))\,.
\end{equation}
The image and exemplars are first split into tokens and concatenated to obtain the token sequence $\bm{F} \in \mathbb{R}^{(L_q + L_e) \times d}$, where $L_q$ and $L_e$ denote the token number of the image and exemplars, respectively, and $d$ denotes the token dimension. A ViT-based encoder with decoupled attention is then applied to extract and to enhance image features. In particular, we use standard ViT-B pretrained with masked autoencoders~\citep{he2022masked} on the FSC-147 dataset as CounTR~\citep{liu2022countr} and CACViT~\citep{wang2024vision}. After feature extraction, we apply a multi-branch box-aware local counter to regress the redundant map, utilizing the scale information from exemplars as a prior to switch among different branches.

\paragraph{Token-level local counter}

Directly applying local counting within ViT could be problematic. In previous CNN-based TasselNets, a local counter regresses counts w.r.t. pixels from arbitrary image patches. Given the input image $\bm I\in\mathbb{R}^{3\times H\times W}$, we extract a set of overlapping image patches by sliding a window of size $k\times k$ with stride $z$, where $k>z$. Each patch is indexed by $i=(i_x,i_y)$, where $i_x \in \{0,...,\lfloor\frac{W-k}{z}\rfloor\}$ and $i_y \in \{0,...,\lfloor\frac{H-k}{z}\rfloor\}$. We then collect the set of image patches within the $i$-th local counting window by
\begin{equation}\label{equ:mm.patch}
\mathcal I_i = \bigl\{\bm I_{z\cdot i + (u,v)}\,\bigm|\,(u,v)\in\{0,\dots,k-1\}^2\,\bigr\}\,.
\end{equation}
After acquiring the set of features $\mathcal{F}_i$ w.r.t. $\mathcal{I}_i$, the redundant count map can be obtained by
\begin{equation}\label{equ:mm.rcount_pixel}
   \bm R = {\underset{i}{\tt cat}}~\mathcal{C}(\mathcal{F}_i)\,.
\end{equation}
Notably, the constraint $k>z$ entails redundant pixels between adjacent patches, formulated by 
\begin{equation}\label{equ:mm.redundant}
\mathcal{I}_i\cap \mathcal{I}_{i'}
=\bigl\{\,I_{z\cdot i'+(u,v)}\mid 0\le N(u,v)<k- z\cdot|i-i'|\,\bigr\}\,,
\end{equation}
where $i$ and $i'$ satisfying
\begin{equation}\label{equ:mm.ii'pixel}
    \|i - i'\|_\infty < \left\lfloor\frac{k}{z} \right\rfloor\,.
\end{equation}
The redundant region characterized by Eq.~\eqref{equ:mm.redundant} is defined at the pixel level, which arises from the pixel-wise overlap between adjacent image patches. We refer the local counter in this case as the pixel-level local counter.

\begin{figure}[!t]
    \centering
    \includegraphics[width=0.8\linewidth]{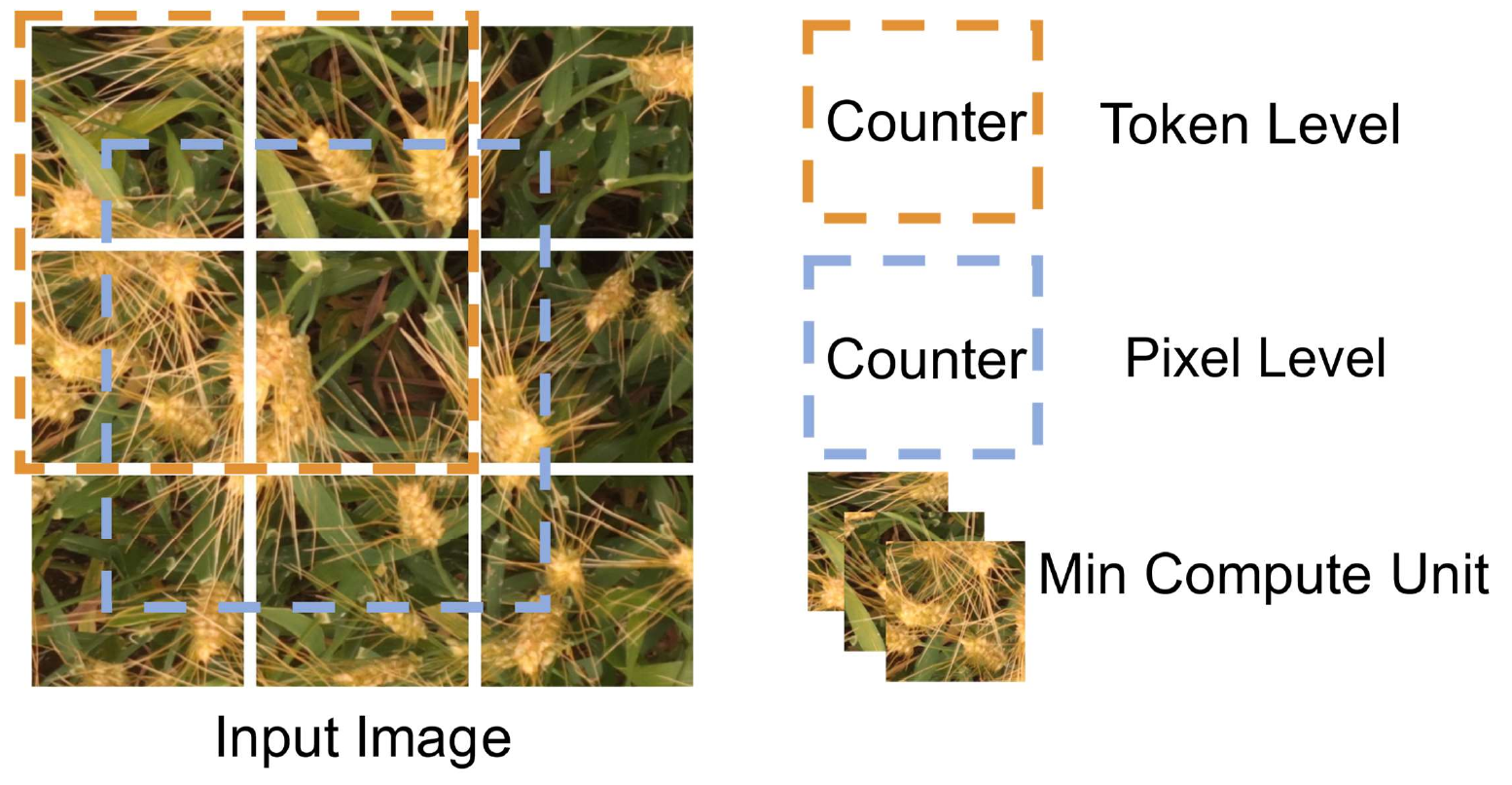}
    \caption{\textbf{How the local counter works at the pixel level and the token level.} The pixel-level local counter counts on arbitrary image patches, while token-level local counter counts on tokens only.}
    \label{fig:mm.redundant}
\end{figure}

Instead of processing image pixels, ViT creates visual representations with image tokens. Tokens are non-overlapped image patches $\mathcal{T}=\{T_{m,n}\in \mathbb{R}^{p\times p} \bigm|\; m\in \{0,...,\frac{H}{p}\},n\in \{0,...,\frac{W}{p}\} \}$, where $p$ denotes the patch size. In this sense, a local counting window used in ViT should contain one or multiple independent tokens. Here we define the set w.r.t. the token-level local counting window of size $k_p \times k_p$ by  
\begin{equation}\label{equ:mm.token window}
\mathcal{T}_j = \bigl\{T_{z\cdot j + (u,v)}\;\bigm|\; (u,v) \in \{0,1,...,k_p-1\}^2\bigr\}\,,
\end{equation}
where $j=(j_x,j_y)$ indexes the center token of the window. 
Thus, we can rewrite Eq.~\eqref{equ:mm.rcount_pixel} by
\begin{equation}\label{equ:token-level counter}
    \bm R = {\underset{j}{\tt cat}}~\mathcal{C}_5(\mathcal{T}_j)\,.
\end{equation}
Following Eq.~\eqref{equ:mm.redundant}, we further define the redundant tokens between adjacent token windows by
\begin{equation}\label{equ:mm.token-level redundancy}
    \mathcal{T}_j\cap \mathcal{T}_{j'}
    =\bigl\{\,T_{z_p\cdot j' + (u,v)}\; \bigm|0\le(u,v)<k_p- z_p\cdot|j-j'| \,\bigr\}\,,
\end{equation}
where $z_p$ is the window stride, and $j$ and $j'$ satisfying 
\begin{equation}\label{equ:mm.ii'}
    \|j - j'\|_\infty < \left\lfloor \frac{k_p}{z_p} \right\rfloor\,.
\end{equation}
In view of the pixel-level local counting window $k$ and the stride $z$, $(k_p,z_p)$ should satisfy 
\begin{equation}\label{equ:mm.integer}
    (k,z) = p\cdot (k_p,z_p)\,.
\end{equation}
For ease of understanding, we use the pixel-level parameters $(k,z)$ in what follows, representing the block size and output stride, respectively, for a unified representation of the local count as in previous TasselNet. Fig.~\ref{fig:mm.redundant} compares the difference between pixel-level and token-level local counting windows. 

The token-level local counter enables the model to count without high-res density maps. This eliminates the need for the ViT decoder, upsampling modules, and subsequent convolutions used in CACViT, simplying the model architecture and enabling faster inference with fewer parameters.

\paragraph{Multi-branch box-aware local counter}

\begin{figure}[!t]
    \centering
    \includegraphics[width=\linewidth]{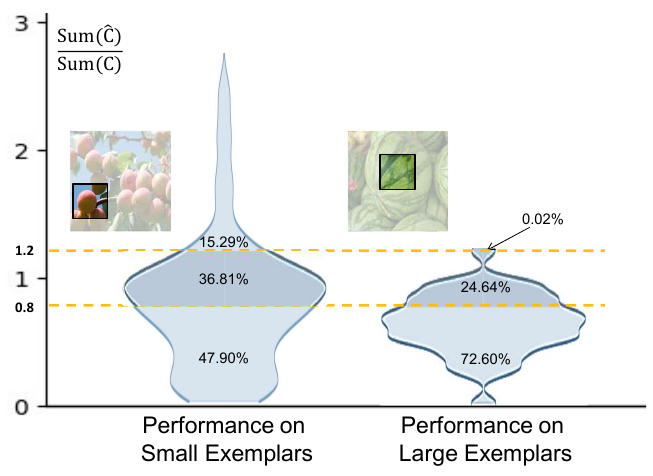}
    \caption{\textbf{The performance distribution of single local counter on the PAC-$105$ dataset with varying exemplar scales.} The vertical axis represents the ratio of the predicted counts to the ground truth. The violin plot on the left represents the performance distribution for exemplars with the size smaller than $32\times 32$, while the one on the right corresponds to exemplars with the size larger than $96 \times 96$. The performance of single local counter reveals a clear distinction between small and large exemplars.}
    \label{fig:mm.premlimitary}
\end{figure}

Albeit effective for non-rigid counting like plants, there is a latent prerequisite behind the working mechanism of the local counting: the block size $k$ should match that of the plant scale. In PAC, plant scales can vary significantly. We presume the single-scale token-level local counter above may be suboptimal to address scale such variations. To verify this, we conduct a sanity check with an experiment on the PAC-$105$. Specifically, we train a single local counter with $k$ set to $32\times32$ and evaluate performance across exemplars of different scales. We consider the exemplars smaller than $32\times32$ pixels as small exemplars, and those larger than $96\times96$ pixels as large ones. We use the ratio of predicted values to ground truth values as the evaluation metric. According to the violin plot in Fig.~\ref{fig:mm.premlimitary}, the single-scale local counter performs better on small exemplars than on large ones. For small exemplars, $36.81\%$ of the samples fall within the interval $[0.8,1.2]$, indicating stable and accurate predictions. However, for large exemplars, only $24.64\%$ of the samples fall within the same interval. Moreover, around $72.60\%$ of the samples fall within the interval $(0,0.8]$, indicating a consistent underestimation on large exemplars. 
We interpret this discrepancy lying in the receptive field of the counter. A sufficiently large receptive field allows the counter to capture 
complete plants entirely; however, a small receptive field may only cover a fraction of the plant, leading to performance degradation. In reality, due to the nature of varying scales in plant sizes and the potential imaging scale fluctuations of different images, it is difficult to choose a local counter with fixed receptive field. 
Since $k$ reflects the receptive field of one local counter, it is rational to set $k$ slightly larger than the size of the counted plant.

To address this challenge, our idea is to design multiple local counters w.r.t.\ different plant scales. During inference, the model can choose an appropriate counter conditioned on the exemplar scale dynamically. We use the exemplar scale $s$ as a prior to estimate the plant scale and represent $s$ as
\begin{equation}\label{equ: scale rule}
    s = \frac{1}{l}\sqrt{\sum_{i=1}^lh_i\sum_{i=1}^lw_i}\,,
\end{equation}
where $h_i,w_i$ denote the height and width of the $i$-th exemplar.

Considering a three-branch local counter, we can rewrite Eq.~\eqref{equ:token-level counter} as
\begin{equation}\label{equ: multihead}
\bm R = 
\begin{cases}
    \mathcal{C}_5^1(\mathcal{E}_4(\bm I, \bm{I}_e,\bm S),\bm S;\theta_1),&\text{if } s \in(0,S_1]\\
    \mathcal{C}_5^2(\mathcal{E}_4(\bm I, \bm{I}_e,\bm S),\bm S;\theta_2),&\text{if } s \in(S_1,S_2]\\
    \mathcal{C}_5^3(\mathcal{E}_4(\bm I, \bm{I}_e,\bm S),\bm S;\theta_3),&\text{if } s \in(S_2,S_3]\\
\end{cases}\,,
\end{equation}
where $\mathcal{C}_5^i$ denotes the $i$-th counter, $\theta_i$ denotes its counter parameters, and ${S}_i$ is a predefined threshold that defines the scale boundary of certain exemplars. Eq.~\eqref{equ: multihead} indicates that the model determines which counter to use conditioned on the interval including the scale prior $s$. We optimize all counters simultaneously during training. Furthermore, an additional slack layer implemented by $3\times 3$ convolution is attached to each counter to adapt features to different plant scales. We will demonstrate the necessity of this slack layer in Sec.~\ref{sec:re.ablation}.

\paragraph{Attention-informed visualizer}
After normalization, we still need a visualizer to inform which regions are counted. 
The previous explainable visualizer in TasselNetV3 is not applicable to TasselNetV4 due to the reliance on high-resolution features. Hence, we propose an attention-informed, parameter-free visualizer specific to TasselNetV4.

\begin{figure}[!t]
    \centering
    \includegraphics[width=\linewidth]
    {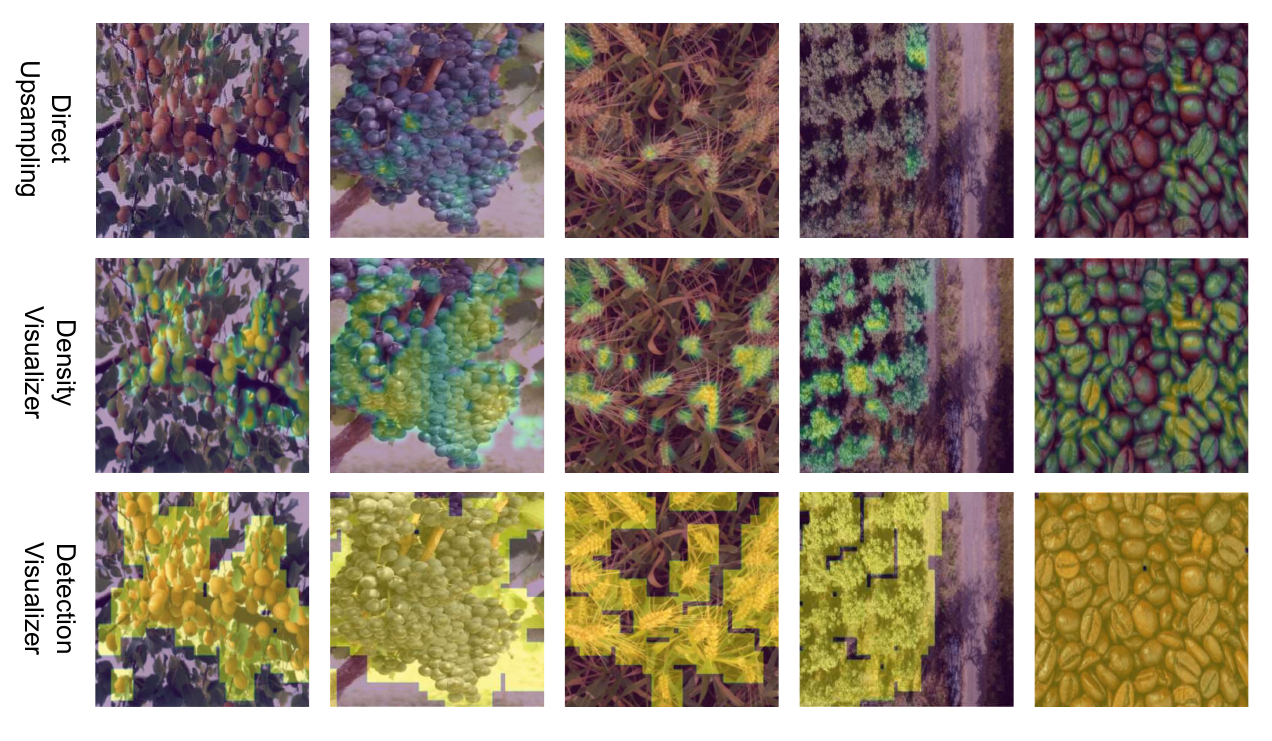}
    \caption{\textbf{Comparison of different visualizers.} One is designed to serve as a density-like representation and the other is designed to serve as a detection-like representation. }
    \label{fig:r.visualizer}
\end{figure}

With the count map $\bm{C} \in \mathbb{R} ^{(\frac{H-b}{o}+1) \times (\frac{W-b}{o}+1)}$ and the attention map $\bm{A}_{match} \in \mathbb{R} ^ {(\frac{H}{p} \times \frac{W}{p})}$ from the last decoupled attention layer, we first interpolate $\bm{C}$ to the same resolution of $\bm{A}_{match}$. Then we select the top-$k$ points in $\bm{A}_{match}$, assign a value of $1$ to these points and $0$ to others to generate a hint map, indicating instance locations. The number of chosen points $N_{{top}}$ is jointly determined by the final predicted count and the magnitude embedding $M_e$ (Eq.~\eqref{equ:magnitude}) as
\begin{equation}\label{equ:topk}
    N_{{top}} = \underset{i,j}{\sum}c_{ij} M_e\,,
\end{equation}
where $c_{ij}$ is the local count in $\bm C$ indexed by $i$ and $j$.
We can tailor the visualizer to two different practical scenarios based on this token-level guidance. 
One is for the detection-like task with these top-$k$ tokens serving as a mask; the other is to apply a Hadamard product between these top-$k$ tokens and the local count map, enabling the visualization of the region distribution and density. As shown in Fig.~\ref{fig:r.visualizer}, both the detection-like visualizer and the density-like visualizer merge similarity and counting information at the token level, thus being aware of the representation and location of the instance.

\paragraph{Training and testing pipelines}
During training, the training images, dotted annotations, up to three exemplars (indicated by bounding boxes) for each image, and the scale embedding of the exemplars are used as inputs to train the model. In the counting head, all-branch counters infer the redundant count map $\bm R$, while only the branch that fits the scale prior $s$ will contribute to backpropagation. We use the standard $\ell_1$ loss, which is defined by
\begin{equation}\label{equ:loss}
    \ell_1 = \sum_{i=1}^{3}\mathds{1}(S_{i-1}<s\leq S_i)\|\bm{R} -\bm R_{gt}\|\,,S_0=0\,,
    \end{equation}
where $\bm R_{gt}$ is the ground-truth redundant count map (cf. TasselNetV2+~\citep{lu2020tasselnetv2+}), and $\mathds{1}(\cdot)$ is the indicator function.

During testing, exemplars are also provided for each testing sample. Different from the training stage, only a single branch counter infers the redundant count map conditioned on the exemplar scale per Eq.~\eqref{equ: multihead}. The same normalizer $\mathcal{N}_2$ as TasselNetV2+ is used to project the predicted redundant map $\bm{\hat{R}}$ back to the normalized count map $\bm{C}$ and obtain the image count by summing over $\bm C$.

\section{Results}\label{sec:result}

Here we present our experimental results. We first introduce the baselines, implementation details, and evaluation metrics. We then present our ablation study to justify our technical improvements. Additional experiments are also conducted to compare plant-specific counting against plant-agnostic counting and to highlight the peculiarity of plants with cross-dataset evaluation on FSC-147~\citep{famnet}.

\subsection{Baselines, implementation details, and evaluation metrics}\label{re.baseline}
\paragraph{Baselines}

We select six CAC approaches as base plant counting models. 

\begin{itemize}
    \item \textbf{FamNet}~\citep{famnet} is a matching-based method for zero-shot counting. It uses a CNN to extract features from images and exemplars and matches them to predict a density map. It also proposes a dataset FSC-$147$, which has become the standard CAC benchmark. 
    \item\textbf{BMNet+}~\citep{shi2022represent} proposes a similarity-aware framework by jointly learning representation and similarity metric in an end-to-end manner. 
    \item \textbf{SAFECount}~\citep{you2023few} extends the similarity-aware design in BMNet+ by introducing similarity-guided feature extraction. 
    \item \textbf{SPDCNet}~\citep{lin2022scale} embeds the scale information into deformable convolution.
    It dynamically adjusts the receptive field according to different instance scales. 
    \item \textbf{CountTR}~\citep{liu2022countr} is one of the first methods introducing ViT into CAC. It regresses density map with an `extract then match' pipeline.
    \item \textbf{T-Rex2}~\citep{jiang2024t} exploits the text-image synergy using CLIP~\citep{radford2021learning} to train an open-world object detection model. This model could be easily adapted to address the task of CAC. 
    \item \textbf{CACViT}~\citep{wang2024vision} introduces the `extract-and-match' pipeline to CAC. It provides a decoupled view on the attention map.  
\end{itemize}


\paragraph{Implementation details}
Here we provide detailed settings of the network structure, data augmentation, and training details.
\begin{itemize}
    \item \textit{Network architecture.}
    TasselNetV4 follows an encoder-only structure. It tokenizes both the input image of size $384 \times384$ and exemplars of size $64\times64$ into patches of size $16 \times 16$, followed by $12$ self-attention layers with $768$ hidden dimensions. At the last self-attention layer, we decouple the attention map. As for the counter, we employ three parallel branches w.r.t. different scales. During training, the block size and output stride $(k,z)$ are set to $(32,16)$, $(64,16)$, and $(128,16)$, respectively. We adopt $3$‑shot and $1$‑shot regimes following the standard CAC evaluation protocol. In the 3‑shot setting, three exemplars are provided during training and testing; in the $1$‑shot setting, a random exemplar is chosen during testing. We iterate over all available exemplars and report the mean and standard deviation.
    
    \item  \textit{Data augmentation.}
    We highlight the random-mosaic method, which merges part of the current image with other images of different plant categories. This appears to be beneficial for similarity modeling. Additionally, we fix the shortest side of the input image to $384$ and resize both the input image and its exemplars while preserving the original aspect ratio. $384\times384$ image patches are cropped as training samples from the resized images.

    \item \textit{Training details.}
    We employ the AdamW optimizer in conjunction with OneCycle scheduling to dynamically adjust the learning rate throughout the training process. The base learning rate is set to $0.0001$. The number of training epochs is set to $200$, and it takes around $6$ hours to converge on a single NVIDIA GeForce RTX $3090$. Notably, we train all models only on PAC-$105$, with PAC-Somalia serving as a test dataset. To improve training efficiency, we apply the following strategies during training:

    \begin{itemize}
        \item \textit{Auto-mixed-precision (amp)}. This technique enables the model to process data with FP-$16$ and FP-$32$ smoothly.
        \item \textit{Flash attention}. This technique optimizes the self-attention mechanism, which significantly reduces memory usage and computation time by computing attention in a fused and hardware-friendly manner. It requires only the precision of FP-$16$.
        \item \textit{Adaptive Gaussian kernel}. We assign the Gaussian kernel size dynamically conditioned on the size of the exemplar(s).
    \end{itemize}
\end{itemize}
To ensure a fair comparison, these adjustments were consistently applied to all baselines.

\paragraph{Evaluation metrics}

We report the standard MAE and RMSE metrics on the testing set. 
WCA and $R^2$ are also provided to assess the practical value, and the Mean Percentage Error (MPE) is introduced to assess over- or under-estimation in the main experiments. These metrics take the form
\begin{equation}
   \mathit{MAE} = \frac{1}{M} \sum_i^M{|p_i - c_i|}\,, 
\end{equation}
\begin{equation}
    \mathit{RMSE} = \sqrt{\frac{1}{M}\sum_i^M(p_i - c_i)^2}\,,
\end{equation}
\begin{equation}
    \mathit{WCA} = 1-\frac{\sum_{i}^M|c_i - p_i|}{\sum_i^M c_i}\,,
\end{equation}
\begin{equation}
    R^2 = 1- \frac{\sum_i^M(p_i-c_i)^2}{\sum_i^M(\bar c-c_i)^2}\,,
\end{equation}
\begin{equation}
    \mathit{MPE} = \frac{1}{M}\sum_i^M\frac{(p_i - c_i)}{c_i + \epsilon}\,,
\end{equation}
where $M$ denotes the number of testing images, $p_i$ indicates the prediction of the $i$-th image, $c_i$ is the corresponding ground truth count, $\bar{c}$ denotes the average ground truth count, and $\epsilon$ denotes a small number. 
Additionally, Frames Per Second (FPS) is reported to quantify the efficiency. 
We report the metrics above under both the $1$-shot and $3$-shot settings (when applicable).

\begin{table*}[t] \small
    \centering
    \caption{\textbf{Comparison with the state-of-the-art CAC approaches on the PAC-$105$ dataset.} Best performance is in \textbf{boldface}.}
    \label{tab:baseline.metrics}
    \renewcommand{\arraystretch}{1.2} 
    \addtolength{\tabcolsep}{0pt} 
    \begin{tabularx}{\textwidth}{@{}lcccccccc@{}}
        \toprule
        
        Method         & Venue \& Year & Shot & MAE$\downarrow$ & RMSE$\downarrow$ & WCA$\uparrow$ & $R^2\uparrow$  & MPE$|\downarrow|$   &FPS$\uparrow$\\
        \midrule 
        FamNet \citep{famnet}            & CVPR'21 & 3 & 31.70  & 62.58  & 0.49   & 0.56   & 0.24
 & 89.65\\ 
        BMNet+ \citep{shi2022represent}  & CVPR'22 & 3 & 27.03  & 60.18  & 0.56   & 0.61   & \textbf{0.02}
 & 43.08\\ 
        SPDCNet \citep{lin2022scale}     & BMVC'22 & 3 & 25.21  & 48.99  & 0.58   & 0.52   & 0.21
 & \textbf{155.84}\\ 
        SAFECount \citep{you2023few}     & WACV'23 & 3 & 25.59  & 52.04  & 0.63   & 0.74   & 0.04
 & 13.98\\ 
        CountTR \citep{liu2022countr}    & BMVC'22 & 3 & 27.65  & 48.24  & 0.62   & 0.76   & 0.12
 & 50.49\\ 
        T-Rex2 \citep{jiang2024t}        & ECCV'24 & 3 & 16.64  & 49.39  & 0.73   & 0.74   & -0.13     & --
        \\ 
        CACViT \citep{wang2024vision}    & AAAI'24 & 3 & 19.51  & 29.59  & 0.68   & 0.89   & 0.29     & 89.65\\ 
        TasselNetV4 (Ours)      & This Paper    & 3 & \textbf{16.04}  & \textbf{28.03}   & \textbf{0.74}  & \textbf{0.92}  & -0.05  & 121.62\\ 
        \midrule
        FamNet \citep{famnet}         & CVPR'21 & 1 & 35.91{\small $\pm0.966$}  & 71.78{\small $\pm1.188$}   & 0.42{\small $\pm0.014$}   & 0.45{\small $\pm0.024$} & 0.57{\small $\pm0.017$} & -- \\  
        BMNet+ \citep{shi2022represent}& CVPR'22 & 1 & 28.78{\small $\pm0.324$}   & 62.12{\small $\pm0.437$}   & 0.15{\small $\pm0.001$}    & 0.59{\small $\pm0.008$} & 0.13{\small $\pm0.006$} & -- \\   
        CountTR \citep{liu2022countr} & BMVC'22 & 1 & 28.46{\small $\pm0.226$}   & 49.84{\small $\pm0.646$}   & 0.70{\small $\pm0.037$}    & 0.73{\small $\pm0.006$} & 0.11{\small $\pm0.003$} & -- \\
        CACViT \citep{wang2024vision} & AAAI'24  & 1 & 21.80{\small $\pm0.429$}  & 38.40{\small $\pm1.526$}  & 0.64{\small $\pm0.005$}    & 0.84{\small $\pm0.013$} & 0.29{\small $\pm0.002$} & -- \\ 
        TasselNetV4 (Ours)      &  This Paper   & 1 & \textbf{18.04}{\small $\pm0.339$}   & \textbf{32.04}{\small $\pm1.213$}    & \textbf{0.71}{\small $\pm0.005$}   & \textbf{0.90}{\small $\pm0.009$}  & \textbf{0.02}{\small $\pm0.000$}  & -- \\
        \bottomrule
    \end{tabularx}
\end{table*}

\begin{table*}[t] \small
    \centering
    \caption{\textbf{Comparison with the state-of-the-art CAC approaches on the PAC-Somalia dataset.} Best performance is in \textbf{boldface}.}
    \label{tab:baseline.metrics:PACS}
    \renewcommand{\arraystretch}{1.2} 
    \addtolength{\tabcolsep}{0pt} 
    \begin{tabularx}{\textwidth}{@{}lccYYYYY@{}}
        \toprule
        Method     & Shot    & Venue \& Year & MAE$\downarrow$ & RMSE$\downarrow$ & WCA$\uparrow$ & $R^2\uparrow$  & MPE$|\downarrow|$  \\
        \midrule 
        CountTR \citep{liu2022countr} & 3 & BMVC'22 & 12.71  & 23.87  & 0.38   & 0.57   & \textbf{0.27}\\ 
        CACViT \citep{wang2024vision} & 3 & AAAI'24  & 14.00  & 17.00  & 0.55   & 0.78  & 0.80\\ 
        TasselNetV4 (Ours)            & 3 & This Paper    & \textbf{8.88}  & \textbf{13.11}  & \textbf{0.72}   & \textbf{0.87}  & 0.32 \\ 
        \midrule
        CountTR \citep{liu2022countr} & 1 & BMVC'22 & 12.79{\small $\pm0.076$}   & 24.20{\small $\pm0.292$}   & 0.37{\small $\pm0.005$}    & 0.55{\small $\pm0.012$}  & \textbf{0.23}{\small $\pm0.062$} \\
        CACViT \citep{wang2024vision} & 1 & AAAI'24  & 14.74{\small $\pm0.138$}  & 18.23{\small $\pm0.446$}  & 0.53{\small $\pm0.005$}    & 0.75{\small $\pm0.012$}  &  0.80{\small $\pm0.008$} \\ 
        TasselNetV4 (Ours)            & 1 &  This Paper   & \textbf{10.98}{\small $\pm0.065$}   & \textbf{16.73}{\small $\pm0.150$}    & \textbf{0.65}{\small $\pm0.000$}   & \textbf{0.80}{\small $\pm0.005$}  & 0.42{\small $\pm0.001$} \\
        \bottomrule
    \end{tabularx}
\end{table*}

\subsection{Main results}
Here we benchmark \OurMethod against several baselines, evaluating the PAC performance on PAC-$105$ and the OOD generalization on PAC-Somalia. 

\begin{figure*}[!t]
    \centering
    \includegraphics[width=\textwidth]
    {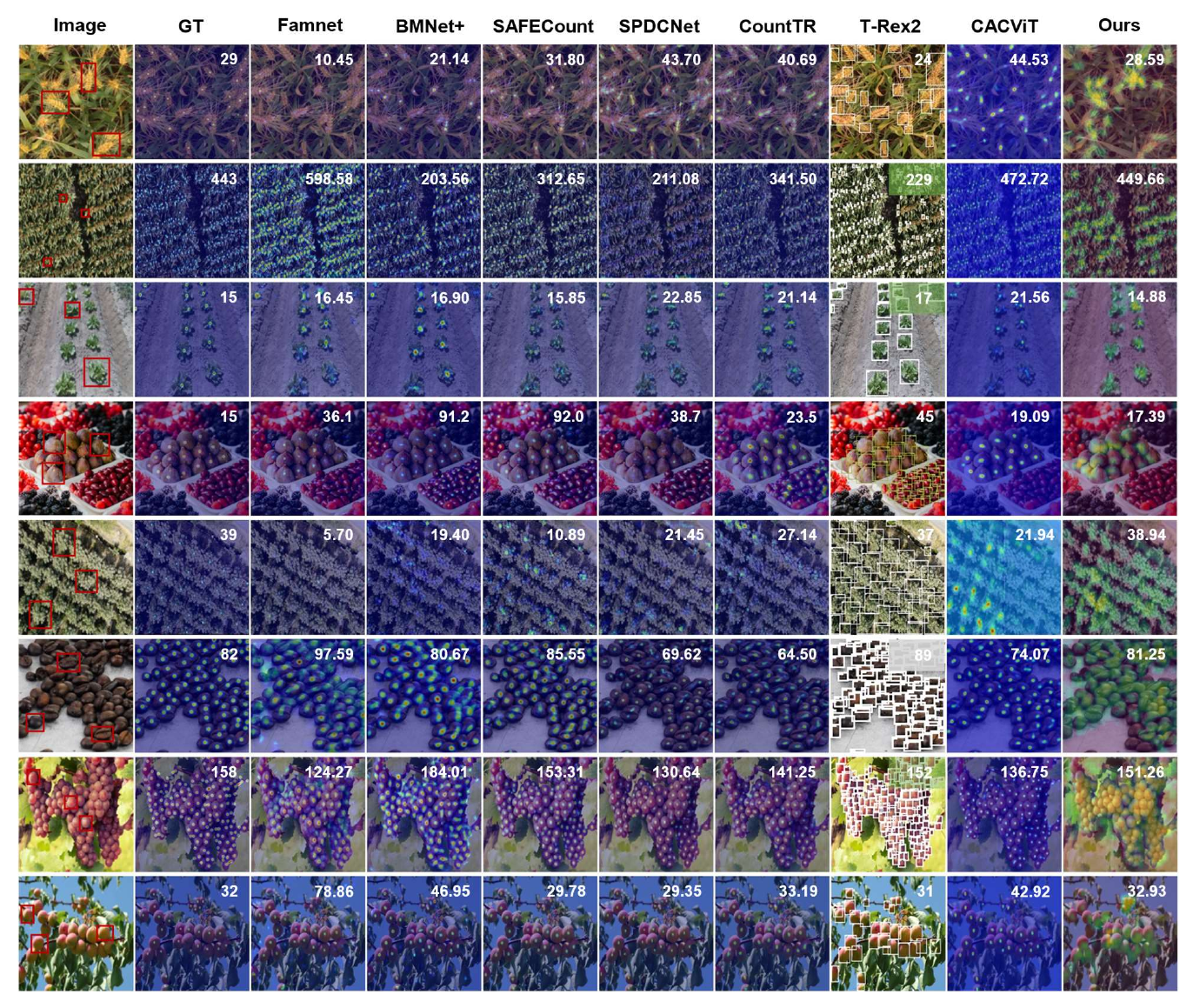}
    \caption{\textbf{Qualitative results of baselines and our method on the PAC-$105$ dataset.} Our method consistently outperforms previous methods with more precise location and counting.}
    \label{fig:r.vis}
\end{figure*}

\paragraph{Results on the PAC-$105$ dataset}
We first evaluate on the PAC-$105$ dataset. 
We compare \OurMethod 
with state-of-the-art CAC methods. 
Results are shown in Table~\ref{tab:baseline.metrics}.
One can see that, generic CAC approaches do not adapt to plants well, even if their models are trained on the PAC-105 dataset, while \OurMethod outperforms them and pushes the state of the art forward under both the $3$-shot and $1$-shot settings, achieving an $R^2$ value of $0.92$ and $0.90$, respectively, significantly outperforming other competitors.
Visualizations of the $3$-shot setting are shown in Fig.~\ref{fig:r.vis}. \OurMethod not only counts more accurately, but also informs individual plants with informative count maps.

\begin{figure}[!t]
    \centering
    \includegraphics[width=\linewidth]
    {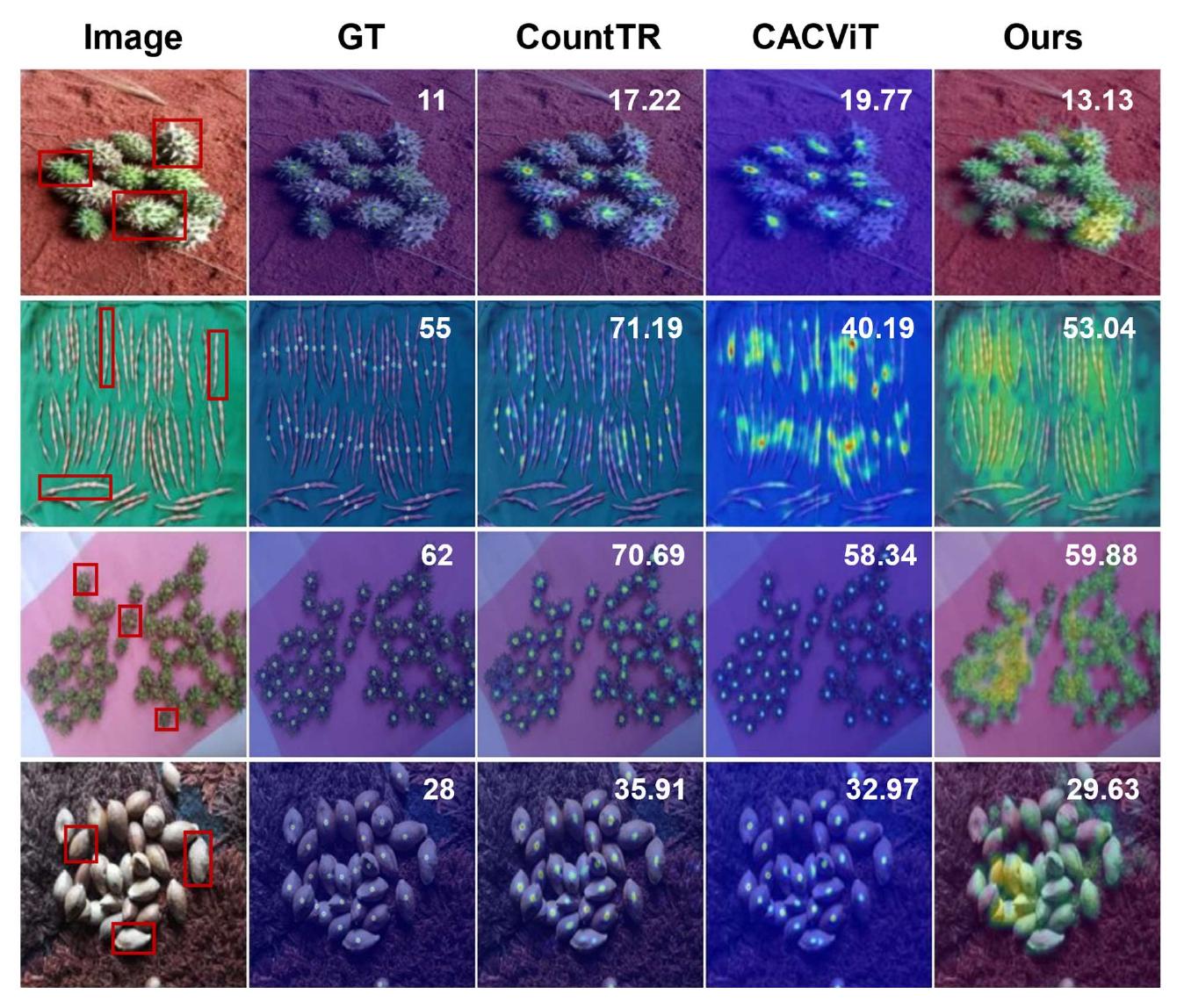}
    \caption{\textbf{Qualitative results of baselines and our method on the PAC-Somalia dataset.} Our method consistently outperforms previous methods with more precise location and counting.}
    \label{fig:r.vis_somalia}
\end{figure}

\paragraph{Results on the PAC-Somalia dataset}
We further evaluate cross-domain generalization by reporting the performance on the PAC-Somalia dataset.
Since PAC-Somalia has no overlapped plant category against PAC-$105$, it is an ideal venue for evaluating the generalization of CAC models trained on PAC-$105$.
For simplicity, we compare \OurMethod with two baselines sharing similar ViT-based architectures.
Results are summarized in Table~\ref{tab:baseline.metrics:PACS}, and visualizations are shown in Fig.~\ref{fig:r.vis_somalia}. 
Among both $3$-shot and $1$-shot settings, \OurMethod consistently achieves superior performance against previous approaches across all metrics.

\begin{table}[!t] \small
    \centering
    \caption{\textbf{Plant-specific counting versus plant-agnostic counting on the MTC dataset.} $^\dagger$ indicates that the model is finetuned on the MTC training set. Best performance is in \textbf{boldface}.}
    \label{tab:mtc}
    \renewcommand{\arraystretch}{1.2} 
    \setlength{\tabcolsep}{7pt}  
    \begin{tabularx}{\columnwidth}{@{}lccc@{}}
        \toprule
        Method      & MAE $\downarrow$ & RMSE $\downarrow$ & $R^2$ $\uparrow$\\
        \midrule 
        MCNN~\citep{zhang2016single}               & 17.9    & 21.9   & 0.33\\
        TasselNet~\citep{lu2017tasselnet}          & 6.6     & 9.9    & 0.87\\
        TasselNetV2~\citep{xiong2019tasselnetv2}   & 5.4     & 9.2    & 0.89\\
        TasselNetV2+~\citep{lu2020tasselnetv2+}    & 5.1     & \textbf{9.0}    & 0.89\\
        TasselNetV3-Lite~\citep{lu2021tasselnetv3} & \textbf{4.9}    & 9.1    & \textbf{0.91}\\
        TasselNetV4 (Ours)       & 11.8    & 18.8    & 0.51\\
        TasselNetV4$^\dagger$ (Ours)       & 5.5    & 10.5    & 0.85\\
        \bottomrule
    \end{tabularx}
\end{table}

\subsection{Plant-agnostic counting versus plant-specific counting}\label{sec:re.mtc}

Here we conduct experiments to answer the question whether plant-agnostic counting models can replace plant-specific ones. We 
compare TasselNetV4 with previous TasselNets~\citep{lu2017tasselnet,lu2020tasselnetv2+,lu2021tasselnetv3} and MCNN~\citep{zhang2016single} on the Maize Tassels Counting (MTC) dataset. In particular, since the PAC-$105$ training set has contained some of the images in the MTC testing set, we remove these images from the PAC-$105$ training set and retrain the model for a fair comparison. Results are shown in Table~\ref{tab:mtc}. We observe that TasselNetV4 falls behind those plant-specific baselines, achieving only $11.8$ in MAE. 
To further justify whether the superiority of previous TasselNets comes from the overfitting of the dataset, we also finetune TasselNetV4 on the MTC training set and report the performance again. Surprisingly, the MAE reduces significantly, from $11.8$ to $5.5$, which suggests existing plant-specific models exhibits a certain degree of overfitting.

\subsection{Cross-dataset evaluation}\label{sec:re.fsc}

We further conduct cross-dataset evaluation to highlight the peculiarity of plants, that is, their non-rigid nature against most rigid daily object classes. 
Specifically, we resort to the FSC-147 dataset~\citep{famnet}, a generic CAC benchmark. 
We compare two settings: i) FSC-$147$$\rightarrow$PAC-$105$ ($F\rightarrow P$), where the model is trained on the FSC-$147$ training set and tested on the PAC-$105$ test set; and ii) PAC-$105$$\rightarrow$FSC-$147$ ($P\rightarrow F$), where the model is trained on the PAC-$105$ training set and tested on the FSC-$147$ test set. 
Additionally, we also compare 
a recent class-agnostic detection counting method GeCo~\citep{pelhan2024novel}. 
Note that GeCo is only applicable to the $F\rightarrow P$ setting due to the unavailability of bounding box annotations in the PAC-$105$ dataset, so we cannot compare it in the main experiments. 
For ease of comparison, we 
report results of in-domain settings as well, that is, FSC-$147$ train $\rightarrow$ FSC-$147$ test ($F\rightarrow F$) and PAC-$105$ train $\rightarrow$ PAC-$105$ test ( $P\rightarrow P$). 
As shown in Table~\ref{tab:fsc}, all methods exhibit significant performance degradation when the training and testing sets differ.
Notably, the $F\rightarrow P$ setting (G2, C2, and T2) yields substantially worse results than the $P\rightarrow F$ setting (C4 and T4).
Even the SOTA method GeCo suffers a huge 
performance drop in the $F\rightarrow P$ setting, with MAE increases from $8.10$ (G1) to $43.49$ (G2). This result underscores that the distinctive, non-rigid characteristics of plants are difficult to learn from predominantly rigid generic object categories, highlighting the need for a dedicated PAC dataset and specially designed models. Optimized for plant counting, TasselNetV4 achieves the best MAE of $16.04$ in the $P\rightarrow P$ setting (T3). 
Interestingly, CACViT serves as a strong CAC baseline on rigid-object counting ($10.83$ MAE in C1 vs. $16.10$ MAE in T1), but fails to perform well in PAC ($19.51$ MAE in C3 vs. $16.04$ MAE in T3); and TasselNetV4 exhibits a reverse tendency, being the best performing model on PAC-105 but slightly falling behind other competitors on FSC-147. This suggests TasselNetV4 indeed has been customized to fit the nature of plants. We will discuss more in detail in Section~\ref{sec:discussion}.

\subsection{Ablation study}\label{sec:re.ablation}

  

\begin{table}[!t]
\centering
\caption{ \textbf{Cross-dataset counting results between PAC-$105$ and FSC-147.} \textit{A} $\rightarrow$ \textit{B} denotes that a model is trained on \textit{A} but tested on \textit{B}.}
\label{tab:fsc}
\setlength{\tabcolsep}{2pt}
\begin{tabularx}{\columnwidth}{@{}lllYYYc@{}}
\toprule
 \multicolumn{3}{l}{Setting}   & \multirow{2}{*}{MAE$\downarrow$}  & \multirow{2}{*}{RMSE$\downarrow$} & \multirow{2}{*}{WCA$\uparrow$}  & \multirow{2}{*}{$R^2\uparrow$} \\
\cmidrule(r){1-3}
 No. & Trained on & Tested on &  &  &  &  \\
\midrule
&\multicolumn{6}{l}{\textit{GeCo}~\citep{pelhan2024novel}} \\
 G1 &FSC-147 $\rightarrow$ &FSC-147   & 8.10   & 60.16  & 0.88  & 0.85     \\  
 G2 &FSC-147 $\rightarrow$ &PAC-$105$   & 43.49  & 142.35 & 0.30  & -1.16    \\
\midrule
&\multicolumn{6}{l}{\textit{CACViT}~\citep{wang2024vision}} \\
 C1 &FSC-147 $\rightarrow$ &FSC-147   & 10.83  & 73.12  & 0.83  & 0.75     \\  
 C2 &FSC-147 $\rightarrow$ &PAC-$105$   & 38.49  & 82.35  & 0.38  & 0.28    \\
 \rowcolor[HTML]{F6F2DF}
 C3 &PAC-$105$ $\rightarrow$ &PAC-$105$   & 19.51  & 29.59  & 0.68  & 0.89     \\
 \rowcolor[HTML]{F6F2DF}
 C4 &PAC-$105$ $\rightarrow$ &FSC-147   & 26.68  & 107.43 & 0.60  & 0.46     \\ 
\midrule
&\multicolumn{6}{l}{\textit{TasselNetV4 (Ours)}} \\
 T1 &FSC-147 $\rightarrow$ &FSC-147   & 16.16  & 110.56 & 0.76  & 0.43     \\  
 T2 &FSC-147 $\rightarrow$ &PAC-$105$   & 51.40  & 123.20 & 0.17  & -0.62    \\
 \rowcolor[HTML]{F6F2DF}
 T3 &PAC-$105$ $\rightarrow$ &PAC-$105$   & 16.04  & 28.03  & 0.74  & 0.92     \\
 \rowcolor[HTML]{F6F2DF}
 T4 &PAC-$105$ $\rightarrow$ &FSC-147   & 25.04  & 122.35 & 0.62  & 0.30     \\ 
\bottomrule
\end{tabularx}
\end{table}

Here we conduct ablation studies to justify the effectiveness of the multi-branch counters, the necessity of the slack layer, and how the local counter enhances model efficiency.

\begin{table*}[!t] \small
    \centering
    \caption{\textbf{Ablation study on the multi-branch box-aware counter.} Best performance is in \textbf{boldface}, and second best is underlined}.
    \label{tab:ablation.multihead}
    \renewcommand{\arraystretch}{1.2} 
    \addtolength{\tabcolsep}{5pt} 
    \begin{tabularx}{\textwidth}{@{}lccccYccccY@{}}
        \toprule
            & &\multicolumn{4}{c}{PAC-$105$}  &&\multicolumn{4}{c}{PAC-Somalia} \\
        \midrule
        Box setting    & Slack layer     & MAE $\downarrow$ & RMSE $\downarrow$ & WCA $\uparrow$ & $R^2$ $\uparrow$ && MAE $\downarrow$ & RMSE $\downarrow$ & WCA $\uparrow$ & $R^2$ $\uparrow$\\
        \midrule  
        $32,128$             & \ding{55} & 23.95& 51.14  & 0.61   & 0.72  & &19.51& 34.95 & 0.38   & 0.08\\
        $32,128$             & \ding{51} & \textbf{15.02}& \textbf{25.24} & \textbf{0.76}  & \textbf{0.93} && \underline{11.73}  & \underline{18.13} & \underline{0.63}   & \underline{0.75} \\
        $32,64,128$          & \ding{51} & \underline{16.04}  & \underline{28.03}  & \underline{0.74}  & \underline{0.92} && \textbf{8.88}  & \textbf{13.11}& \textbf{0.72}   & \textbf{0.87}\\
        \bottomrule
    \end{tabularx}
\end{table*}

\paragraph{Multi-branch box-aware local counters}

\begin{table}[!t] \small
    \centering
    \caption{\textbf{PAC-$105$ results stratified by small exemplars ($<32\times 32$) and large exemplars ($>96\times 96$).} Best performance is in \textbf{boldface}, and second best is \underline{underline}.}
    \label{tab:ablation.trend}
    \renewcommand{\arraystretch}{1.2} 
    \addtolength{\tabcolsep}{-1pt} 
    \begin{tabularx}{\columnwidth}{@{}lcccccc@{}}
        \toprule
            &\multicolumn{3}{c}{Small exemplar set}  &\multicolumn{3}{c}{Large exemplar set} \\
        \midrule
        Box setting      & MAE $\downarrow$ & RMSE $\downarrow$ & $R^2$ $\uparrow$ & MAE $\downarrow$ & RMSE $\downarrow$ & $R^2$ $\uparrow$ \\
        \midrule  
        $32$         & \underline{24.94}  & \textbf{35.64}   & \textbf{0.93}    & 5.36  & 6.89   & -0.05\\
        $64$         & 31.37  & 48.28    & 0.87  & 4.47  & 5.82  & 0.25\\
        $80$         & 25.11  & 40.68    & 0.90  & 4.43  & 5.83  & 0.25\\
        $96$         & 30.32  & 53.05    & 0.84  & 3.75  & 4.80  & 0.49\\
        $112$        & \textbf{24.66}  & \underline{38.73}   & 0.91    & 3.83  & 4.88 & 0.48\\
        $128$        & 25.55  & 41.56    & 0.90  & \textbf{3.65} & \underline{4.73}  & \underline{0.51}\\
        $144$        & 29.54  & 47.27    & 0.87  & 6.98  & 8.78  & -0.70 \\
        $160$        & 30.37  & 45.05    & 0.87  & 5.73  & 7.42  & -0.22\\
        $192$        & 35.18  & 54.48    & 0.83  & 5.81  & 7.52  & -0.25\\
        \midrule
        $32, 64, 128$ & 26.01  & 39.41  & \underline{0.91}  & \underline{3.73}  & \textbf{4.61}  & \textbf{0.53}\\
        \bottomrule
    \end{tabularx}
\end{table}

\begin{figure*}[!t]
    \centering
    \includegraphics[width=\textwidth]
    {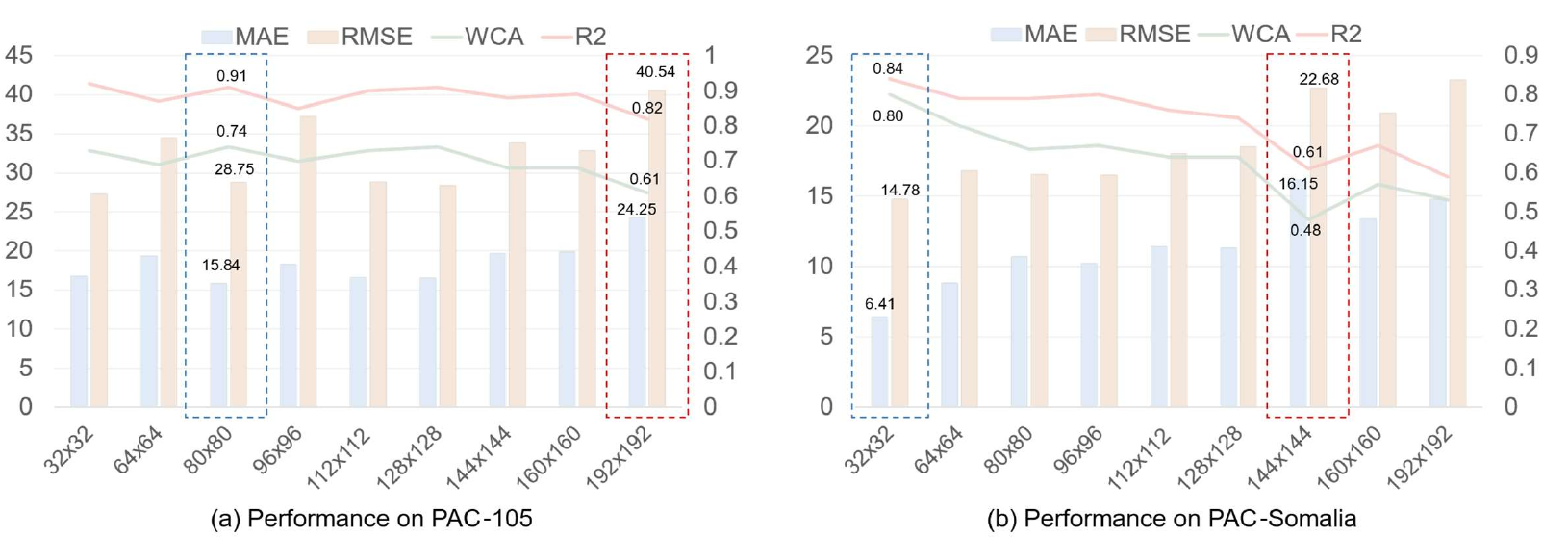}
    \caption{\textbf{Effect of the block size $k$.} (a) Performance on the PAC-$105$ dataset. (b) Performance on the PAC-Somalia dataset. Best results are in blue, and worst results are in red.}
    \label{fig:r.ablation}
\end{figure*}

We first justify the design of multi-brach box-aware counters. We begin with training several single-scale counters of the model and study their generalization across PAC-$105$ and PAC-Somalia. Fig.~\ref{fig:r.ablation} illustrates how performance changes with changed block sizes. 
Across PAC-$105$ and PAC-Somalia, we observe a general tendency 
of decreased performance with increased block sizes. Specifically, on PAC-$105$, there are cases where counters with 
rather different block sizes perform similarly, such as $32\times 32$ and $128\times 128$. We conjecture that the effective block size 
may correlate with the exemplar size. 
To validate this hypothesis, we partition the test set of PAC-$105$ into two subsets following the setting of Fig.~\ref{fig:mm.premlimitary}: 
i) a small-exemplar subset with exemplar sizes smaller than $32\times 32$ and ii) a large-exemplar subset with sizes larger than $96\times 96$. We re-evaluate those single-scale counters quantitatively. As shown in Table~\ref{tab:ablation.trend}, for single-branch counters in large exemplars, the performance of MAE and R$^2$ improves when the block size increases, reaches the peak at $128\times 128$, and then degrades for larger ones. 
For small exemplars, we observe the 
performance fluctuates with different block sizes, with $32\times 32$ yielding the best $R^2$ and the $80\times80$ and $112\times112$ working competitively. 
Such size-conditioned behaviors explain why $32\times 32$ and $128\times 128$ produce comparable performance on the PAC-$105$ test set: $32\times 32$ fits small exemplars but fails to tackle the large ones, and the opposite for $128\times128$.

We then investigate the effectiveness of the slack layer. Table~\ref{tab:ablation.multihead} reveals a significant performance improvement, suggesting the slack layer plays a pivotal role in aligning the general features learned by the backbone with the specific requirements of the different box-aware counters. 
Furthermore, a joint analysis of Table~\ref{tab:ablation.multihead},~\ref{tab:ablation.trend}, and Fig.~\ref{fig:r.ablation} demonstrates that the incorporation of the multi-branch architecture contributes to more stable performance across different PAC-$105$ and PAC-Somalia. 

\paragraph{Efficiency improvement}

Here we study to what extent the local counter could improve the efficiency of the model. 
We compare our TasselNetV4 with a modified CACViT to make sure that different performance is caused \textit{only by the local counter}. 
We compute Floating Point Operations Per Second (FLOPs) and the number of parameters. 
According to Table~\ref{tab:ablation.efficiency}, compared with CACViT, the use of local counters reduces both learnable and non-learnable parameters. Meanwhile, TasselNetV4 exhibits significantly lower GFLOPs, making it 
more friendly 
to resource-constrained scenarios.

\section{Discussions}
\label{sec:discussion}

\paragraph{TasselNetV4 emerges to be an effective vision foundation model for cross-scene, cross-scale, and cross-species plant counting} 
While previous TasselNets have been developed and optimized for plant counting, they are tailored to specific plant species and scenes and cannot generalize across species. Inspired by recent work on CAC, TasselNetV4 inherits the vein of its ancestors but takes a great leap towards PAC. With versatile PAC datasets and specially designed scale-aware counters, \OurMethod can act like a foundation model and achieves cross-scene, cross-scale, and cross-species plant counting, without retraining on new species.


\paragraph{Plant-agnostic counting is fine-grained class-agnostic counting} Unlike common daily objects studied in computer vision, plants are non-rigid, dynamic in space and time, and exhibit large intra‑variations, which violates the common assumption in CAC that intra-class variations are smaller than inter-class ones. This insight is supported by our cross-dataset results. According to Table~\ref{tab:fsc}, transferring a CAC model from FSC-$147$ to PAC-$105$ (F$\rightarrow$P) yields consistently poor performance, indicating that the representations learned from common rigid objects can struggle to 
generalize to non-rigid plants. 
These observations motivate us to treat PAC as a new task and to design customized models taking the fine-grained characteristics of plants into account.


\begin{table}[!t] \small
    \centering
    \caption{\textbf{Comparison of computational overhead.}}
    \label{tab:ablation.efficiency}
    \renewcommand{\arraystretch}{1.2} 
    \addtolength{\tabcolsep}{7pt} 
    \begin{tabularx}{\columnwidth}{@{}lccc@{}}
        \toprule
        \multirow{2}{*}{methods}        & \multicolumn{2}{c}{\#Param.} &  \multirow{2}{*}{GFLOPs$\downarrow$ }   \\
        \cline{2-3}
        &Learnable   &Non-learnable &   \\ \hline
        CACViT              & 99.77M   & 0.75M  & 97.30\\ 
        TasselNetV4         & 84.67M   & 0.45M  & 52.37\\ 
        \bottomrule
    \end{tabularx}
\end{table}

\paragraph{Transferring from non-rigid plants to rigid generic objects appears easier than the opposite} 
While the cross-dataset results in Table~\ref{tab:fsc} have confirmed the soundness of PAC, we also observe another interesting phenomenon: $P\rightarrow F$ shows less performance degradation than $F\rightarrow P$ (cf. (C2--C1) vs.\ (C4--C3) and (T2--T1) vs.\ (T4--T3), where (A--B) indicates the metric difference between A and B). This suggests the representations learned from non-rigid plants seem to transfer more easily to rigid objects. 
This is consistent with the intuition that a model trained on difficult cases often generalizes to simpler ones, whereas the converse does not necessarily hold. 


\paragraph{\OurMethod is customized and optimized for plant-agnostic counting}
According to Table~\ref{tab:fsc}, compared with the standard CAC baselines, \OurMethod underperforms them on FSC-147 (cf. G1, C1, and T1), but outperforms CACViT on PAC-105 (cf. T3 vs. C3). Such results imply that \OurMethod has been customized and optimized to the plant domain, at the cost of sacrificing certain capability to perceiving generic object categories.
This may stem from the local counting paradigm adopted by \OurMethod.
Compared with prior CAC methods that rely on density map regression which often under‑represents deformed objects, token-level local counting can yield better robustness to non-rigid deformations. It is worth noting that \OurMethod does not overfit the plant domain, because it still maintains non-trivial class-agnostic transferability in the setting of $P\rightarrow F$ (cf. T4 vs. C4).

\paragraph{Plant-agnostic counting points out a promising direction for plant counting with ample scope for future improvement} 
This insight is backed by the plant-specific counting results on the MTC dataset~\citep{lu2017tasselnet} in Table~\ref{tab:mtc}. According to the results, directly applying pretrained \OurMethod to MTC yields suboptimal performance. While a simple finetuning somehow fixes the problem, the results suggest there still exists a clear performance gap between plant-agnostic counting and plant-specific counting. The increased sensibility to broad categories seems to come at the cost of decreased counting accuracy. We remark that this phenomenon, while being undesirable, is in accord with the famous \textit{no free lunch} theorem, saying that ``if an algorithm performs well on a certain class of problems then it necessarily pays for that with degraded performance on the set of all remaining problems''~\citep{wolpert2002no}.
Nevertheless, on the positive side, the performance gap also reveals a goal to develop better PAC approaches that are on par with or even outperform plant-specific models.

\paragraph{Token-level local counting is more effective than density-based regression for plants} 
Previous work has shown that pixel-level local counting outperforms density map regression~\citep{lu2017tasselnet}. 
Yet, pixel-level local counting cannot be directly applied to ViT.
In this work, we generalize the pixel-level local counter to a token-level formulation and demonstrate its effectiveness for plant counting, particularly in plant-agnostic settings. 
Our experiments show consistent performance gains.
This advantage stems from the inherent robustness of local counting.
In density-map regression, each instance is encoded with a Gaussian kernel. 
This encoding is often inadequate for capturing non-rigid plant deformations.
By contrast, local counting only requires the target plant to fall within a local window, yielding a fixed count regardless of shape or deformation.
Taken together, these results show that the token-level local counter well suits plant counting and is compatible with both convolutional networks and ViTs.

\paragraph{Local counting improves computational efficiency}
Local counting also simplifies the architecture because it does not require high-resolution outputs.
As shown in Table~\ref{tab:ablation.efficiency}, our method reduces parameters by $15.13\%$ and GFLOPs by $46.18\%$ relative to CACViT, indicating improved computational efficiency. 
This improvement primarily stems from the token-level local counter, which eliminates the decoder in CACViT.
In CACViT, the decoder comprises four multi-head self-attention layers followed by a convolutional density regressor, yielding $O(N^2)$ time complexity because of global self-attention.
By contrast, our model applies a local counter directly to tokens, achieving a lower time complexity of $O(N \times k^2)$.
This simplification substantially facilitates deployment, making the model more computationally efficient than CACViT on resource-constrained platforms (e.g., UAVs).

\paragraph{Multi-branch box-aware local counters enhance cross-scale and cross-species generalization} 
This observation is supported by a comparison of the single local counter with the multi-branch, box-aware local counter.
As shown in Fig.~\ref{fig:r.ablation}, the single local counter has two main limitations.
First, the optimal block size setting varies substantially between PAC-$105$ and PAC-Somalia. 
The best setting of the single local counter on the PAC-$105$ yields poor performance on the PAC-Somalia, and vice versa. 
This suggests limited generalization to OOD species. 
Second, the single-scale local counter is sensitive to the parameter of block size, which stems from the relationship between the exemplar size and the block size. As shown in Table~\ref{tab:ablation.trend}, single-scale counters of different block size have different optimal intervals w.r.t. exemplar sizes. A small block size single-scale counter can fit well for plants with small exemplar size, but struggles to handle plants with large-sized exemplars. For a single-scale counter of large block size, vice versa. 
In contrast, the multi-branch, box-aware local counter aggregates these advantageous intervals, achieving competitive performance on both subsets, and delivers stable cross-scale plant counting.  
Though demonstrating effective and efficient, the multi-branch box-aware local counters still have limitations in design. Our current hard switching is heuristic, where branch selection relies on an exemplar-derived scale prior with a preset threshold. Table~\ref{tab:ablation.trend} further reveals that the optimal block size is not always the one slightly larger than the exemplar, which is likely due to that the exemplar size is an imperfect proxy for the distribution of instance sizes within the image. All these suggest promising directions for our future work.

\section{Conclusion}
In this work, we introduce plant-agnostic counting and build the first two datasets specific to this task, i.e., PAC-$105$ and PAC-Somalia. Technically, we present \OurMethod, an effective, efficient approach based on a naive, plain ViT architecture. We also customize this ViT-based framework with the local counting idea by extending the previous pixel-level local counter to the token-level one. To enhance scale robustness, we introduce the multi-branch box-aware local counter, which leverages multiple parallel counters and a hard switching mechanism based on the exemplar scale. Extensive experiments on the two datasets demonstrate that \OurMethod enables cross-scene, cross-scale, and cross-species plant counting. 

For future work, we plan to enrich the datasets with more plant species and extend the application scenarios to broader observation scales such as microscopy, broadening the applicability of plant counting to a wider range of real-world scenarios. 

\section*{Acknowledgement}
This work was partly supported by the National Natural Science Foundation of China under Grant No.~62576146, partly supported by the National Key Research and Development Program of China under Grant No.~2022YFE0204700, No.~2022YFD2300700, and No.~2022YFE0116200, and partly by the Young Scientists Fund of the National Natural Science Foundation of China under Grant No.~42201437. Wei Guo's participation was partly supported by the Sarabetsu Village ``Endowed Chair for Field Phenomics'' project in Hokkaido and partly by JSPS KAKENHI Grant No.~JP25H01110, Japan.

\bibliographystyle{elsarticle-harv}
\bibliography{reference}

\begin{thebibliography}{52}
\expandafter\ifx\csname natexlab\endcsname\relax\def\natexlab#1{#1}\fi
\providecommand{\url}[1]{\texttt{#1}}
\providecommand{\href}[2]{#2}
\providecommand{\path}[1]{#1}
\providecommand{\DOIprefix}{doi:}
\providecommand{\ArXivprefix}{arXiv:}
\providecommand{\URLprefix}{URL: }
\providecommand{\Pubmedprefix}{pmid:}
\providecommand{\doi}[1]{\href{http://dx.doi.org/#1}{\path{#1}}}
\providecommand{\Pubmed}[1]{\href{pmid:#1}{\path{#1}}}
\providecommand{\bibinfo}[2]{#2}
\ifx\xfnm\relax \def\xfnm[#1]{\unskip,\space#1}\fi
\bibitem[{Ak{\c{c}}ay et~al.(2020)Ak{\c{c}}ay, Kabasakal, Aksu, Demir, {\"O}z and Erdo{\u{g}}an}]{akccay2020automated}
\bibinfo{author}{Ak{\c{c}}ay, H.G.}, \bibinfo{author}{Kabasakal, B.}, \bibinfo{author}{Aksu, D.}, \bibinfo{author}{Demir, N.}, \bibinfo{author}{{\"O}z, M.}, \bibinfo{author}{Erdo{\u{g}}an, A.}, \bibinfo{year}{2020}.
\newblock \bibinfo{title}{Automated bird counting with deep learning for regional bird distribution mapping}.
\newblock \bibinfo{journal}{Animals} \bibinfo{volume}{10}, \bibinfo{pages}{1207}.
\bibitem[{Bai et~al.(2023)Bai, Liu, Cao, Lu, Xiong, Yang, Cai, Wang and Yao}]{bai2023rice}
\bibinfo{author}{Bai, X.}, \bibinfo{author}{Liu, P.}, \bibinfo{author}{Cao, Z.}, \bibinfo{author}{Lu, H.}, \bibinfo{author}{Xiong, H.}, \bibinfo{author}{Yang, A.}, \bibinfo{author}{Cai, Z.}, \bibinfo{author}{Wang, J.}, \bibinfo{author}{Yao, J.}, \bibinfo{year}{2023}.
\newblock \bibinfo{title}{Rice plant counting, locating, and sizing method based on high-throughput uav rgb images}.
\newblock \bibinfo{journal}{Plant Phenomics} \bibinfo{volume}{5}, \bibinfo{pages}{0020}.
\bibitem[{Bargoti and Underwood(2017)}]{bargoti2017deep}
\bibinfo{author}{Bargoti, S.}, \bibinfo{author}{Underwood, J.}, \bibinfo{year}{2017}.
\newblock \bibinfo{title}{Deep fruit detection in orchards}, in: \bibinfo{booktitle}{Int. Conf. Robot. Autom.}, \bibinfo{organization}{IEEE}. pp. \bibinfo{pages}{3626--3633}.
\bibitem[{Buzzy et~al.(2020)Buzzy, Thesma, Davoodi and Mohammadpour~Velni}]{buzzy2020real}
\bibinfo{author}{Buzzy, M.}, \bibinfo{author}{Thesma, V.}, \bibinfo{author}{Davoodi, M.}, \bibinfo{author}{Mohammadpour~Velni, J.}, \bibinfo{year}{2020}.
\newblock \bibinfo{title}{Real-time plant leaf counting using deep object detection networks}.
\newblock \bibinfo{journal}{Sensors} \bibinfo{volume}{20}, \bibinfo{pages}{6896}.
\bibitem[{Chamara et~al.(2023)Chamara, Bai and Ge}]{chamara2023aicropcam}
\bibinfo{author}{Chamara, N.}, \bibinfo{author}{Bai, G.}, \bibinfo{author}{Ge, Y.}, \bibinfo{year}{2023}.
\newblock \bibinfo{title}{Aicropcam: Deploying classification, segmentation, detection, and counting deep-learning models for crop monitoring on the edge}.
\newblock \bibinfo{journal}{Comput. Electron. Agric.} \bibinfo{volume}{215}, \bibinfo{pages}{108420}.
\bibitem[{David et~al.(2020)David, Madec, Sadeghi-Tehran, Aasen, Zheng, Liu, Kirchgessner, Ishikawa, Nagasawa, Badhon et~al.}]{david2020global}
\bibinfo{author}{David, E.}, \bibinfo{author}{Madec, S.}, \bibinfo{author}{Sadeghi-Tehran, P.}, \bibinfo{author}{Aasen, H.}, \bibinfo{author}{Zheng, B.}, \bibinfo{author}{Liu, S.}, \bibinfo{author}{Kirchgessner, N.}, \bibinfo{author}{Ishikawa, G.}, \bibinfo{author}{Nagasawa, K.}, \bibinfo{author}{Badhon, M.A.}, et~al., \bibinfo{year}{2020}.
\newblock \bibinfo{title}{Global wheat head detection (gwhd) dataset: A large and diverse dataset of high-resolution rgb-labelled images to develop and benchmark wheat head detection methods}.
\newblock \bibinfo{journal}{Plant Phenomics} .
\bibitem[{Deng et~al.(2021)Deng, Tao, Huang, Bangura, Jiang, Jiang and Qi}]{deng2021automated}
\bibinfo{author}{Deng, R.}, \bibinfo{author}{Tao, M.}, \bibinfo{author}{Huang, X.}, \bibinfo{author}{Bangura, K.}, \bibinfo{author}{Jiang, Q.}, \bibinfo{author}{Jiang, Y.}, \bibinfo{author}{Qi, L.}, \bibinfo{year}{2021}.
\newblock \bibinfo{title}{Automated counting grains on the rice panicle based on deep learning method}.
\newblock \bibinfo{journal}{Sensors} \bibinfo{volume}{21}, \bibinfo{pages}{281}.
\bibitem[{Dosovitskiy et~al.(2020)Dosovitskiy, Beyer, Kolesnikov, Weissenborn, Zhai, Unterthiner, Dehghani, Minderer, Heigold, Gelly et~al.}]{dosovitskiy2020image}
\bibinfo{author}{Dosovitskiy, A.}, \bibinfo{author}{Beyer, L.}, \bibinfo{author}{Kolesnikov, A.}, \bibinfo{author}{Weissenborn, D.}, \bibinfo{author}{Zhai, X.}, \bibinfo{author}{Unterthiner, T.}, \bibinfo{author}{Dehghani, M.}, \bibinfo{author}{Minderer, M.}, \bibinfo{author}{Heigold, G.}, \bibinfo{author}{Gelly, S.}, et~al., \bibinfo{year}{2020}.
\newblock \bibinfo{title}{An image is worth 16x16 words: Transformers for image recognition at scale}.
\newblock \bibinfo{journal}{arXiv Comput. Res. Repository} .
\bibitem[{Ghosal et~al.(2019)Ghosal, Zheng, Chapman, Potgieter, Jordan, Wang, Singh, Singh, Hirafuji, Ninomiya et~al.}]{ghosal2019weakly}
\bibinfo{author}{Ghosal, S.}, \bibinfo{author}{Zheng, B.}, \bibinfo{author}{Chapman, S.C.}, \bibinfo{author}{Potgieter, A.B.}, \bibinfo{author}{Jordan, D.R.}, \bibinfo{author}{Wang, X.}, \bibinfo{author}{Singh, A.K.}, \bibinfo{author}{Singh, A.}, \bibinfo{author}{Hirafuji, M.}, \bibinfo{author}{Ninomiya, S.}, et~al., \bibinfo{year}{2019}.
\newblock \bibinfo{title}{A weakly supervised deep learning framework for sorghum head detection and counting}.
\newblock \bibinfo{journal}{Plant Phenomics} .
\bibitem[{Giuffrida et~al.(2016)Giuffrida, Minervini and Tsaftaris}]{giuffrida2016learning}
\bibinfo{author}{Giuffrida, M.V.}, \bibinfo{author}{Minervini, M.}, \bibinfo{author}{Tsaftaris, S.A.}, \bibinfo{year}{2016}.
\newblock \bibinfo{title}{Learning to count leaves in rosette plants}.
\newblock \bibinfo{journal}{Proc. Comput. Vis. Probl. Plant Phenotyping Workshop} .
\bibitem[{Group(2016)}]{VGG2023VIA}
\bibinfo{author}{Group, V.G.}, \bibinfo{year}{2016}.
\newblock \bibinfo{title}{Vgg image annotator : a standalone image annotator application}.
\newblock \bibinfo{howpublished}{\url{https://gitlab.com/vgg/via}}.
\newblock \bibinfo{note}{Accessed: 2025-09-10}.
\bibitem[{Guo et~al.(2018)Guo, Zheng, Potgieter, Diot, Watanabe, Noshita, Jordan, Wang, Watson, Ninomiya et~al.}]{guo2018aerial}
\bibinfo{author}{Guo, W.}, \bibinfo{author}{Zheng, B.}, \bibinfo{author}{Potgieter, A.B.}, \bibinfo{author}{Diot, J.}, \bibinfo{author}{Watanabe, K.}, \bibinfo{author}{Noshita, K.}, \bibinfo{author}{Jordan, D.R.}, \bibinfo{author}{Wang, X.}, \bibinfo{author}{Watson, J.}, \bibinfo{author}{Ninomiya, S.}, et~al., \bibinfo{year}{2018}.
\newblock \bibinfo{title}{Aerial imagery analysis--quantifying appearance and number of sorghum heads for applications in breeding and agronomy}.
\newblock \bibinfo{journal}{Front. Plant Sci.} \bibinfo{volume}{9}, \bibinfo{pages}{1544}.
\bibitem[{Guo et~al.(2022)Guo, Wu, Du and Zhang}]{guo2022density}
\bibinfo{author}{Guo, Y.}, \bibinfo{author}{Wu, C.}, \bibinfo{author}{Du, B.}, \bibinfo{author}{Zhang, L.}, \bibinfo{year}{2022}.
\newblock \bibinfo{title}{Density map-based vehicle counting in remote sensing images with limited resolution}.
\newblock \bibinfo{journal}{ISPRS J. Photogramm. Remote Sens.} \bibinfo{volume}{189}, \bibinfo{pages}{201--217}.
\bibitem[{Hasan et~al.(2018)Hasan, Chopin, Laga and Miklavcic}]{hasan2018detection}
\bibinfo{author}{Hasan, M.M.}, \bibinfo{author}{Chopin, J.P.}, \bibinfo{author}{Laga, H.}, \bibinfo{author}{Miklavcic, S.J.}, \bibinfo{year}{2018}.
\newblock \bibinfo{title}{Detection and analysis of wheat spikes using convolutional neural networks}.
\newblock \bibinfo{journal}{Plant Methods} \bibinfo{volume}{14}, \bibinfo{pages}{1--13}.
\bibitem[{He et~al.(2022)He, Chen, Xie, Li, Doll{\'a}r and Girshick}]{he2022masked}
\bibinfo{author}{He, K.}, \bibinfo{author}{Chen, X.}, \bibinfo{author}{Xie, S.}, \bibinfo{author}{Li, Y.}, \bibinfo{author}{Doll{\'a}r, P.}, \bibinfo{author}{Girshick, R.}, \bibinfo{year}{2022}.
\newblock \bibinfo{title}{Masked autoencoders are scalable vision learners}, in: \bibinfo{booktitle}{{IEEE} Conf. Comput. Vis. Pattern Recognit.}, pp. \bibinfo{pages}{16000--16009}.
\bibitem[{Itzhaky et~al.(2018)Itzhaky, Farjon, Khoroshevsky, Shpigler and Bar-Hillel}]{itzhaky2018leaf}
\bibinfo{author}{Itzhaky, Y.}, \bibinfo{author}{Farjon, G.}, \bibinfo{author}{Khoroshevsky, F.}, \bibinfo{author}{Shpigler, A.}, \bibinfo{author}{Bar-Hillel, A.}, \bibinfo{year}{2018}.
\newblock \bibinfo{title}{Leaf counting: Multiple scale regression and detection using deep cnns.}, in: \bibinfo{booktitle}{Brit. Mach. Vis. Conf.}, \bibinfo{organization}{Newcastle}.
\bibitem[{Jiang et~al.(2024)Jiang, Li, Zeng, Ren, Liu and Zhang}]{jiang2024t}
\bibinfo{author}{Jiang, Q.}, \bibinfo{author}{Li, F.}, \bibinfo{author}{Zeng, Z.}, \bibinfo{author}{Ren, T.}, \bibinfo{author}{Liu, S.}, \bibinfo{author}{Zhang, L.}, \bibinfo{year}{2024}.
\newblock \bibinfo{title}{T-rex2: Towards generic object detection via text-visual prompt synergy}, in: \bibinfo{booktitle}{Eur. Conf. Comput. Vis.}, \bibinfo{organization}{Springer}. pp. \bibinfo{pages}{38--57}.
\bibitem[{Khoroshevsky et~al.(2021)Khoroshevsky, Khoroshevsky and Bar-Hillel}]{khoroshevsky2021parts}
\bibinfo{author}{Khoroshevsky, F.}, \bibinfo{author}{Khoroshevsky, S.}, \bibinfo{author}{Bar-Hillel, A.}, \bibinfo{year}{2021}.
\newblock \bibinfo{title}{Parts-per-object count in agricultural images: Solving phenotyping problems via a single deep neural network}.
\newblock \bibinfo{journal}{Remote Sens.} \bibinfo{volume}{13}, \bibinfo{pages}{2496}.
\bibitem[{Kitano et~al.(2019)Kitano, Mendes, Geus, Oliveira and Souza}]{kitano2019corn}
\bibinfo{author}{Kitano, B.T.}, \bibinfo{author}{Mendes, C.C.}, \bibinfo{author}{Geus, A.R.}, \bibinfo{author}{Oliveira, H.C.}, \bibinfo{author}{Souza, J.R.}, \bibinfo{year}{2019}.
\newblock \bibinfo{title}{Corn plant counting using deep learning and uav images}.
\newblock \bibinfo{journal}{{IEEE} Trans. Geosci. Remote Sens.} .
\bibitem[{Lempitsky and Zisserman(2010)}]{NIPS2010_fe73f687}
\bibinfo{author}{Lempitsky, V.}, \bibinfo{author}{Zisserman, A.}, \bibinfo{year}{2010}.
\newblock \bibinfo{title}{Learning to count objects in images}, in: \bibinfo{editor}{Lafferty, J.}, \bibinfo{editor}{Williams, C.}, \bibinfo{editor}{Shawe-Taylor, J.}, \bibinfo{editor}{Zemel, R.}, \bibinfo{editor}{Culotta, A.} (Eds.), \bibinfo{booktitle}{Adv. Neural Inf. Proc. Syst.}, \bibinfo{publisher}{Curran Associates, Inc.}
\newblock \URLprefix \url{https://proceedings.neurips.cc/paper_files/paper/2010/file/fe73f687e5bc5280214e0486b273a5f9-Paper.pdf}.
\bibitem[{Li et~al.(2023)Li, Zheng, Li, Long, Li and Gao}]{li2023tomato}
\bibinfo{author}{Li, P.}, \bibinfo{author}{Zheng, J.}, \bibinfo{author}{Li, P.}, \bibinfo{author}{Long, H.}, \bibinfo{author}{Li, M.}, \bibinfo{author}{Gao, L.}, \bibinfo{year}{2023}.
\newblock \bibinfo{title}{Tomato maturity detection and counting model based on mhsa-yolov8}.
\newblock \bibinfo{journal}{Sensors} \bibinfo{volume}{23}, \bibinfo{pages}{6701}.
\bibitem[{Lin et~al.(2020)Lin, Zhou, Chen, Yu, Wu, Ge, Liu, Li, Jiang, Tang et~al.}]{lin2020heterosis}
\bibinfo{author}{Lin, T.}, \bibinfo{author}{Zhou, C.}, \bibinfo{author}{Chen, G.}, \bibinfo{author}{Yu, J.}, \bibinfo{author}{Wu, W.}, \bibinfo{author}{Ge, Y.}, \bibinfo{author}{Liu, X.}, \bibinfo{author}{Li, J.}, \bibinfo{author}{Jiang, X.}, \bibinfo{author}{Tang, W.}, et~al., \bibinfo{year}{2020}.
\newblock \bibinfo{title}{Heterosis-associated genes confer high yield in super hybrid rice}.
\newblock \bibinfo{journal}{Theor. Appl. Genet.} \bibinfo{volume}{133}, \bibinfo{pages}{3287--3297}.
\bibitem[{Lin et~al.(2014)Lin, Maire, Belongie, Hays, Perona, Ramanan, Doll{\'a}r and Zitnick}]{lin2014microsoft}
\bibinfo{author}{Lin, T.Y.}, \bibinfo{author}{Maire, M.}, \bibinfo{author}{Belongie, S.}, \bibinfo{author}{Hays, J.}, \bibinfo{author}{Perona, P.}, \bibinfo{author}{Ramanan, D.}, \bibinfo{author}{Doll{\'a}r, P.}, \bibinfo{author}{Zitnick, C.L.}, \bibinfo{year}{2014}.
\newblock \bibinfo{title}{Microsoft coco: Common objects in context}, in: \bibinfo{booktitle}{Eur. Conf. Comput. Vis.}, \bibinfo{organization}{Springer}. pp. \bibinfo{pages}{740--755}.
\bibitem[{Lin et~al.(2022)Lin, Yang, Ma, Gao, Liu, Liu, Hou, Yi and Chan}]{lin2022scale}
\bibinfo{author}{Lin, W.}, \bibinfo{author}{Yang, K.}, \bibinfo{author}{Ma, X.}, \bibinfo{author}{Gao, J.}, \bibinfo{author}{Liu, L.}, \bibinfo{author}{Liu, S.}, \bibinfo{author}{Hou, J.}, \bibinfo{author}{Yi, S.}, \bibinfo{author}{Chan, A.B.}, \bibinfo{year}{2022}.
\newblock \bibinfo{title}{Scale-prior deformable convolution for exemplar-guided class-agnostic counting.}, in: \bibinfo{booktitle}{Brit. Mach. Vis. Conf.}, p. \bibinfo{pages}{313}.
\bibitem[{Liu et~al.(2023)Liu, Lu, Cao and Liu}]{liu2023point}
\bibinfo{author}{Liu, C.}, \bibinfo{author}{Lu, H.}, \bibinfo{author}{Cao, Z.}, \bibinfo{author}{Liu, T.}, \bibinfo{year}{2023}.
\newblock \bibinfo{title}{Point-query quadtree for crowd counting, localization, and more}, in: \bibinfo{booktitle}{{IEEE} Int. Conf. Comput. Vis.}, pp. \bibinfo{pages}{1676--1685}.
\bibitem[{Liu et~al.(2022)Liu, Zhong, Zisserman and Xie}]{liu2022countr}
\bibinfo{author}{Liu, C.}, \bibinfo{author}{Zhong, Y.}, \bibinfo{author}{Zisserman, A.}, \bibinfo{author}{Xie, W.}, \bibinfo{year}{2022}.
\newblock \bibinfo{title}{Countr: Transformer-based generalised visual counting}.
\newblock \bibinfo{journal}{Brit. Mach. Vis. Conf.} .
\bibitem[{Lu et~al.(2024)Lu, Nnadozie, Camenzind, Hu and Yu}]{lu2024maize}
\bibinfo{author}{Lu, C.}, \bibinfo{author}{Nnadozie, E.}, \bibinfo{author}{Camenzind, M.P.}, \bibinfo{author}{Hu, Y.}, \bibinfo{author}{Yu, K.}, \bibinfo{year}{2024}.
\newblock \bibinfo{title}{Maize plant detection using uav-based rgb imaging and yolov5}.
\newblock \bibinfo{journal}{Front. Plant Sci.} \bibinfo{volume}{14}, \bibinfo{pages}{1274813}.
\bibitem[{Lu et~al.(2023)Lu, Ye, Wang and Yu}]{lu2023plant}
\bibinfo{author}{Lu, D.}, \bibinfo{author}{Ye, J.}, \bibinfo{author}{Wang, Y.}, \bibinfo{author}{Yu, Z.}, \bibinfo{year}{2023}.
\newblock \bibinfo{title}{Plant detection and counting: Enhancing precision agriculture in uav and general scenes}.
\newblock \bibinfo{journal}{IEEE Access} .
\bibitem[{Lu et~al.(2018)Lu, Xie and Zisserman}]{lu2018class}
\bibinfo{author}{Lu, E.}, \bibinfo{author}{Xie, W.}, \bibinfo{author}{Zisserman, A.}, \bibinfo{year}{2018}.
\newblock \bibinfo{title}{Class-agnostic counting}, in: \bibinfo{booktitle}{Proc. Asian Conf. Comput. Vis.}, \bibinfo{organization}{Springer}. pp. \bibinfo{pages}{669--684}.
\bibitem[{Lu and Cao(2020)}]{lu2020tasselnetv2+}
\bibinfo{author}{Lu, H.}, \bibinfo{author}{Cao, Z.}, \bibinfo{year}{2020}.
\newblock \bibinfo{title}{Tasselnetv2+: A fast implementation for high-throughput plant counting from high-resolution rgb imagery}.
\newblock \bibinfo{journal}{Front. Plant Sci.} \bibinfo{volume}{11}, \bibinfo{pages}{541960}.
\bibitem[{Lu et~al.(2017)Lu, Cao, Xiao, Zhuang and Shen}]{lu2017tasselnet}
\bibinfo{author}{Lu, H.}, \bibinfo{author}{Cao, Z.}, \bibinfo{author}{Xiao, Y.}, \bibinfo{author}{Zhuang, B.}, \bibinfo{author}{Shen, C.}, \bibinfo{year}{2017}.
\newblock \bibinfo{title}{Tasselnet: counting maize tassels in the wild via local counts regression network}.
\newblock \bibinfo{journal}{Plant Methods} \bibinfo{volume}{13}, \bibinfo{pages}{1--17}.
\bibitem[{Lu et~al.(2021)Lu, Liu, Li, Zhao, Wang and Cao}]{lu2021tasselnetv3}
\bibinfo{author}{Lu, H.}, \bibinfo{author}{Liu, L.}, \bibinfo{author}{Li, Y.N.}, \bibinfo{author}{Zhao, X.M.}, \bibinfo{author}{Wang, X.Q.}, \bibinfo{author}{Cao, Z.G.}, \bibinfo{year}{2021}.
\newblock \bibinfo{title}{Tasselnetv3: Explainable plant counting with guided upsampling and background suppression}.
\newblock \bibinfo{journal}{{IEEE} Trans. Geosci. Remote Sens.} \bibinfo{volume}{60}, \bibinfo{pages}{1--15}.
\bibitem[{Mundhenk et~al.(2016)Mundhenk, Konjevod, Sakla and Boakye}]{mundhenk2016large}
\bibinfo{author}{Mundhenk, T.N.}, \bibinfo{author}{Konjevod, G.}, \bibinfo{author}{Sakla, W.A.}, \bibinfo{author}{Boakye, K.}, \bibinfo{year}{2016}.
\newblock \bibinfo{title}{A large contextual dataset for classification, detection and counting of cars with deep learning}, in: \bibinfo{booktitle}{Eur. Conf. Comput. Vis.}, \bibinfo{organization}{Springer}. pp. \bibinfo{pages}{785--800}.
\bibitem[{Pelhan et~al.(2024)Pelhan, Lukezic, Zavrtanik and Kristan}]{pelhan2024novel}
\bibinfo{author}{Pelhan, J.}, \bibinfo{author}{Lukezic, A.}, \bibinfo{author}{Zavrtanik, V.}, \bibinfo{author}{Kristan, M.}, \bibinfo{year}{2024}.
\newblock \bibinfo{title}{A novel unified architecture for low-shot counting by detection and segmentation}.
\newblock \bibinfo{journal}{Adv. Neural Inf. Proc. Syst.} \bibinfo{volume}{37}, \bibinfo{pages}{66260--66282}.
\bibitem[{Radford et~al.(2021)Radford, Kim, Hallacy, Ramesh, Goh, Agarwal, Sastry, Askell, Mishkin, Clark et~al.}]{radford2021learning}
\bibinfo{author}{Radford, A.}, \bibinfo{author}{Kim, J.W.}, \bibinfo{author}{Hallacy, C.}, \bibinfo{author}{Ramesh, A.}, \bibinfo{author}{Goh, G.}, \bibinfo{author}{Agarwal, S.}, \bibinfo{author}{Sastry, G.}, \bibinfo{author}{Askell, A.}, \bibinfo{author}{Mishkin, P.}, \bibinfo{author}{Clark, J.}, et~al., \bibinfo{year}{2021}.
\newblock \bibinfo{title}{Learning transferable visual models from natural language supervision}, in: \bibinfo{booktitle}{Int. Conf. Mach. Learn.}, \bibinfo{organization}{PMLR}. pp. \bibinfo{pages}{8748--8763}.
\bibitem[{Rahnemoonfar and Sheppard(2017)}]{rahnemoonfar2017deep}
\bibinfo{author}{Rahnemoonfar, M.}, \bibinfo{author}{Sheppard, C.}, \bibinfo{year}{2017}.
\newblock \bibinfo{title}{Deep count: fruit counting based on deep simulated learning}.
\newblock \bibinfo{journal}{Sensors} \bibinfo{volume}{17}, \bibinfo{pages}{905}.
\bibitem[{Ranjan et~al.(2021)Ranjan, Sharma, Nguyen and Hoai}]{famnet}
\bibinfo{author}{Ranjan, V.}, \bibinfo{author}{Sharma, U.}, \bibinfo{author}{Nguyen, T.}, \bibinfo{author}{Hoai, M.}, \bibinfo{year}{2021}.
\newblock \bibinfo{title}{Learning to count everything}, in: \bibinfo{booktitle}{{IEEE} Conf. Comput. Vis. Pattern Recognit.}, pp. \bibinfo{pages}{3393--3402}.
\newblock \DOIprefix\doi{10.1109/CVPR46437.2021.00340}.
\bibitem[{Redmon et~al.(2016)Redmon, Divvala, Girshick and Farhadi}]{redmon2016you}
\bibinfo{author}{Redmon, J.}, \bibinfo{author}{Divvala, S.}, \bibinfo{author}{Girshick, R.}, \bibinfo{author}{Farhadi, A.}, \bibinfo{year}{2016}.
\newblock \bibinfo{title}{You only look once: Unified, real-time object detection}, in: \bibinfo{booktitle}{{IEEE} Conf. Comput. Vis. Pattern Recognit.}, pp. \bibinfo{pages}{779--788}.
\bibitem[{Shi et~al.(2022)Shi, Lu, Feng, Liu and Cao}]{shi2022represent}
\bibinfo{author}{Shi, M.}, \bibinfo{author}{Lu, H.}, \bibinfo{author}{Feng, C.}, \bibinfo{author}{Liu, C.}, \bibinfo{author}{Cao, Z.}, \bibinfo{year}{2022}.
\newblock \bibinfo{title}{Represent, compare, and learn: A similarity-aware framework for class-agnostic counting}, in: \bibinfo{booktitle}{{IEEE} Conf. Comput. Vis. Pattern Recognit.}, pp. \bibinfo{pages}{9529--9538}.
\bibitem[{Valente et~al.(2020)Valente, Sari, Kooistra, Kramer and M{\"u}cher}]{valente2020automated}
\bibinfo{author}{Valente, J.}, \bibinfo{author}{Sari, B.}, \bibinfo{author}{Kooistra, L.}, \bibinfo{author}{Kramer, H.}, \bibinfo{author}{M{\"u}cher, S.}, \bibinfo{year}{2020}.
\newblock \bibinfo{title}{Automated crop plant counting from very high-resolution aerial imagery}.
\newblock \bibinfo{journal}{Precis. Agric.} \bibinfo{volume}{21}, \bibinfo{pages}{1366--1384}.
\bibitem[{Vaswani(2017)}]{vaswani2017attention}
\bibinfo{author}{Vaswani, A.}, \bibinfo{year}{2017}.
\newblock \bibinfo{title}{Attention is all you need}.
\newblock \bibinfo{journal}{Adv. Neural Inf. Proc. Syst.} .
\bibitem[{Wang et~al.(2024a)Wang, Yang, Wang, Liang and Chen}]{wang2024satcount}
\bibinfo{author}{Wang, Y.}, \bibinfo{author}{Yang, B.}, \bibinfo{author}{Wang, X.}, \bibinfo{author}{Liang, C.}, \bibinfo{author}{Chen, J.}, \bibinfo{year}{2024}a.
\newblock \bibinfo{title}{Satcount: A scale-aware transformer-based class-agnostic counting framework}.
\newblock \bibinfo{journal}{Neural Netw.} \bibinfo{volume}{172}, \bibinfo{pages}{106126}.
\bibitem[{Wang et~al.(2024b)Wang, Xiao, Cao and Lu}]{wang2024vision}
\bibinfo{author}{Wang, Z.}, \bibinfo{author}{Xiao, L.}, \bibinfo{author}{Cao, Z.}, \bibinfo{author}{Lu, H.}, \bibinfo{year}{2024}b.
\newblock \bibinfo{title}{Vision transformer off-the-shelf: A surprising baseline for few-shot class-agnostic counting}, in: \bibinfo{booktitle}{AAAI Conf. Artif. Intell.}, pp. \bibinfo{pages}{5832--5840}.
\bibitem[{Wolpert and Macready(2002)}]{wolpert2002no}
\bibinfo{author}{Wolpert, D.H.}, \bibinfo{author}{Macready, W.G.}, \bibinfo{year}{2002}.
\newblock \bibinfo{title}{No free lunch theorems for optimization}.
\newblock \bibinfo{journal}{{IEEE} Trans. Evol. Comput.} \bibinfo{volume}{1}, \bibinfo{pages}{67--82}.
\bibitem[{Xiong et~al.(2019)Xiong, Cao, Lu, Madec, Liu and Shen}]{xiong2019tasselnetv2}
\bibinfo{author}{Xiong, H.}, \bibinfo{author}{Cao, Z.}, \bibinfo{author}{Lu, H.}, \bibinfo{author}{Madec, S.}, \bibinfo{author}{Liu, L.}, \bibinfo{author}{Shen, C.}, \bibinfo{year}{2019}.
\newblock \bibinfo{title}{Tasselnetv2: in-field counting of wheat spikes with context-augmented local regression networks}.
\newblock \bibinfo{journal}{Plant methods} \bibinfo{volume}{15}, \bibinfo{pages}{150}.
\bibitem[{Xue et~al.(2016)Xue, Ray, Hugh and Bigras}]{xue2016cell}
\bibinfo{author}{Xue, Y.}, \bibinfo{author}{Ray, N.}, \bibinfo{author}{Hugh, J.}, \bibinfo{author}{Bigras, G.}, \bibinfo{year}{2016}.
\newblock \bibinfo{title}{Cell counting by regression using convolutional neural network}, in: \bibinfo{booktitle}{Eur. Conf. Comput. Vis.}, \bibinfo{organization}{Springer}. pp. \bibinfo{pages}{274--290}.
\bibitem[{Yang et~al.(2021)Yang, Gao, Gao and Zhu}]{yang2021rapid}
\bibinfo{author}{Yang, B.}, \bibinfo{author}{Gao, Z.}, \bibinfo{author}{Gao, Y.}, \bibinfo{author}{Zhu, Y.}, \bibinfo{year}{2021}.
\newblock \bibinfo{title}{Rapid detection and counting of wheat ears in the field using yolov4 with attention module}.
\newblock \bibinfo{journal}{Agronomy} \bibinfo{volume}{11}, \bibinfo{pages}{1202}.
\bibitem[{You et~al.(2023)You, Yang, Luo, Lu, Cui and Le}]{you2023few}
\bibinfo{author}{You, Z.}, \bibinfo{author}{Yang, K.}, \bibinfo{author}{Luo, W.}, \bibinfo{author}{Lu, X.}, \bibinfo{author}{Cui, L.}, \bibinfo{author}{Le, X.}, \bibinfo{year}{2023}.
\newblock \bibinfo{title}{Few-shot object counting with similarity-aware feature enhancement}, in: \bibinfo{booktitle}{Proc. IEEE/CVF Winter Conf. Appl. Comput. Vis.}, pp. \bibinfo{pages}{6315--6324}.
\bibitem[{Zhang et~al.(2015)Zhang, Li, Wang and Yang}]{zhang2015cross}
\bibinfo{author}{Zhang, C.}, \bibinfo{author}{Li, H.}, \bibinfo{author}{Wang, X.}, \bibinfo{author}{Yang, X.}, \bibinfo{year}{2015}.
\newblock \bibinfo{title}{Cross-scene crowd counting via deep convolutional neural networks}, in: \bibinfo{booktitle}{{IEEE} Conf. Comput. Vis. Pattern Recognit.}, pp. \bibinfo{pages}{833--841}.
\bibitem[{Zhang et~al.(2016)Zhang, Zhou, Chen, Gao and Ma}]{zhang2016single}
\bibinfo{author}{Zhang, Y.}, \bibinfo{author}{Zhou, D.}, \bibinfo{author}{Chen, S.}, \bibinfo{author}{Gao, S.}, \bibinfo{author}{Ma, Y.}, \bibinfo{year}{2016}.
\newblock \bibinfo{title}{Single-image crowd counting via multi-column convolutional neural network}, in: \bibinfo{booktitle}{{IEEE} Conf. Comput. Vis. Pattern Recognit.}, pp. \bibinfo{pages}{589--597}.
\bibitem[{Zhu et~al.(2021)Zhu, Hu, Mao, Li, Li, Zhao, Luo, Liu and Yuan}]{zhu2021deep}
\bibinfo{author}{Zhu, C.}, \bibinfo{author}{Hu, Y.}, \bibinfo{author}{Mao, H.}, \bibinfo{author}{Li, S.}, \bibinfo{author}{Li, F.}, \bibinfo{author}{Zhao, C.}, \bibinfo{author}{Luo, L.}, \bibinfo{author}{Liu, W.}, \bibinfo{author}{Yuan, X.}, \bibinfo{year}{2021}.
\newblock \bibinfo{title}{A deep learning-based method for automatic assessment of stomatal index in wheat microscopic images of leaf epidermis}.
\newblock \bibinfo{journal}{Front. Plant Sci.} \bibinfo{volume}{12}, \bibinfo{pages}{716784}.
\bibitem[{Zou et~al.(2020)Zou, Lu, Li, Liu and Cao}]{zou2020maize}
\bibinfo{author}{Zou, H.}, \bibinfo{author}{Lu, H.}, \bibinfo{author}{Li, Y.}, \bibinfo{author}{Liu, L.}, \bibinfo{author}{Cao, Z.}, \bibinfo{year}{2020}.
\newblock \bibinfo{title}{Maize tassels detection: A benchmark of the state of the art}.
\newblock \bibinfo{journal}{Plant Methods} \bibinfo{volume}{16}, \bibinfo{pages}{108}.

\end{thebibliography}

\newpage
\end{document}